\documentclass[letterpaper]{article} 
\usepackage{aaai24}  
\usepackage{times}  
\usepackage{helvet}  
\usepackage{courier}  
\usepackage[hyphens]{url}  
\usepackage{graphicx} 
\usepackage{natbib}  
\usepackage{caption} 
\usepackage{algorithm}
\usepackage{algorithmic}

\usepackage{newfloat}
\usepackage{listings}
\DeclareCaptionStyle{ruled}{labelfont=normalfont,labelsep=colon,strut=off} 
\floatstyle{ruled}
\newfloat{listing}{tb}{lst}{}
\floatname{listing}{Listing}
\usepackage{graphicx,subcaption}
\usepackage{amssymb, amsmath}
\usepackage{multirow}
\usepackage{makecell}
\usepackage{colortbl}
\usepackage[dvipsnames,table]{xcolor}
\usepackage{hhline}
\usepackage[short]{optidef}

\DeclareMathOperator*{\argmax}{argmax}

\title{SafeAR: Safe Algorithmic Recourse by Risk-Aware Policies}

\author {
    Haochen Wu\textsuperscript{\rm 1},
    Shubham Sharma\textsuperscript{\rm 2},
    Sunandita Patra\textsuperscript{\rm 2},
    Sriram Gopalakrishnan\textsuperscript{\rm 2}
}
\affiliations {
    \textsuperscript{\rm 1} University of Michigan, Ann Arbor\\
    \textsuperscript{\rm 2} J.P. Morgan AI Research\\
    haochenw@umich.edu,
    shubham.x2.sharma@jpmchase.com,
    sunandita.patra@jpmchase.com,
    sriram.gopalakrishnan@jpmchase.com
}

\nocopyright
\usepackage{tikz}

\begin{document}
\maketitle

\renewcommand{\arraystretch}{1.2}

\begin{tikzpicture}[remember picture, overlay]
  \node [text=blue, yshift = 2cm, font=\Large] at (current page.south)  {Accepted to the main track of AAAI 2024 with an oral presentation};
\end{tikzpicture}

\begin{abstract}
With the growing use of machine learning (ML) models in critical domains such as finance and healthcare, the need to offer recourse for those adversely affected by the decisions of ML models has become more important; individuals ought to be provided with recommendations on actions to take for improving their situation and thus receiving a favorable decision. Prior work on sequential algorithmic recourse---which recommends a series of changes---focuses on action feasibility and uses the proximity of feature changes to determine action costs. However, the uncertainties of feature changes and the risk of higher than average costs in recourse have not been considered. It is undesirable if a recourse could (with some probability) result in a worse situation from which recovery requires an extremely high cost. It is essential to incorporate risks when computing and evaluating recourse. We call the recourse computed with such risk considerations as Safe Algorithmic Recourse (SafeAR). The objective is to empower people to choose a recourse based on their risk tolerance. In this work, we discuss and show how existing recourse desiderata can fail to capture the risk of higher costs. We present a method to compute recourse policies that consider variability in cost and connect algorithmic recourse literature with risk-sensitive reinforcement learning. We also adopt measures ``Value at Risk'' and ``Conditional Value at Risk'' from the financial literature to summarize risk concisely. We apply our method to two real-world datasets and compare policies with different risk-aversion levels using risk measures and recourse desiderata (sparsity and proximity).
\end{abstract}

\section*{Introduction}

\begin{figure}[!t]
    \centering
    \includegraphics[width=0.47\textwidth]{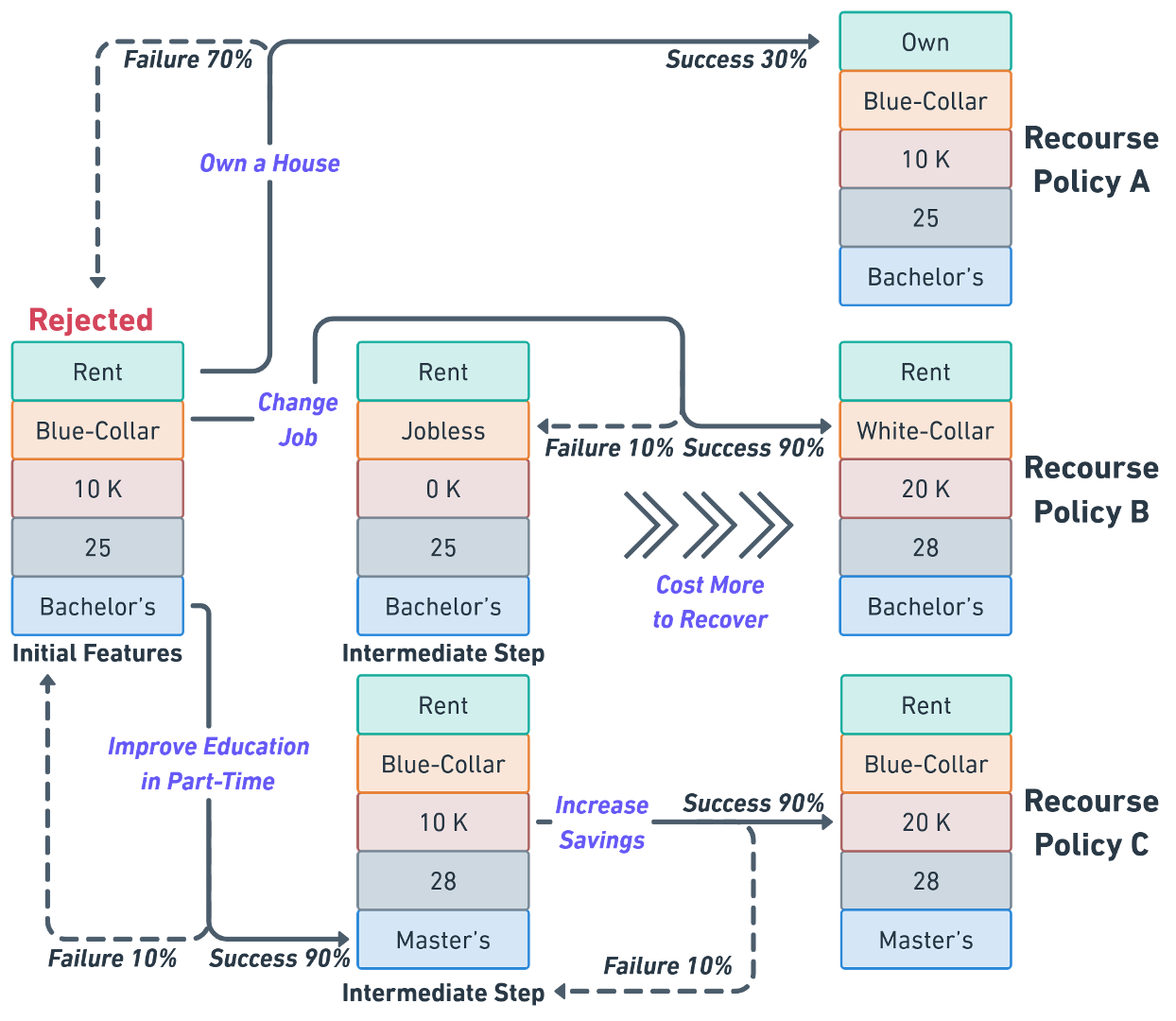}
    \caption{Recourse policies for loan approvals. Policy A has only one feature change (low sparsity) but with a high failure rate; Policy B has the lowest expected cost but might result in a situation that costs more to recover; Policy C has a slightly higher expected cost than Policy B but lower variance in cost (risk), which can be considered as a safer Policy.}
    \label{fig:stable_plan}
\end{figure}

Machine learning (ML) models are increasingly being used to make decisions in a wide array of scenarios including healthcare~\cite{bigdata_ml_healthcare}, insurance premiums \cite{ML_for_health_insurance}, and loan approvals~\cite{ml_loan_approval}. Given the impact of ML models on society, the importance of algorithmic recourse has increased ~\cite{venkatasubramanian2020philosophical}. Algorithmic recourse refers to a computed recommendation provided to an end user that suggests specific changes they can make to convert an unfavorable outcome (e.g., loan rejection), into a favorable one. For a recourse to be helpful, the suggested change ought to be actionable; for example, one can change their savings balance but not their age. Existing recourse work has considered the cost of taking the recommended actions~\cite{SCM_costs_causal_models,AmortizedRL,FACE}. However, they do not consider the risk of higher costs. In this work, \emph{risk} means the potential for higher costs during the recourse due to the possibility of reaching adverse states; this can happen due to uncertainties in action effects (not deterministic). The cost could be in terms of time required, effort, financial resources, etc. Without incorporating risks---which is ignoring uncertainties or only minimizing the expected costs---the recipient of an algorithmic recourse may be caught unaware and unprepared for situations with high costs. By offering recourse policies with risk measures, we can help people be aware of how much risk is involved in each policy and choose a safer one. The \emph{recourse policy} here refers to the recommended actions for all possible states a person might encounter, as opposed to a single deterministic sequence of actions.

To further understand the need for risk considerations, let's look at the existing algorithmic recourse approaches that use \emph{counterfactual explanation} (CE) methods to give recourse recommendations. CE methods find ``the most similar instances to the feature vector describing the individual, that result in the desired prediction from the model'' \cite{Review4}. The assumption is that minimizing feature-space differences translates to a recourse that requires less cost to reach the desired outcome. Rather than providing a single vector of feature changes, recourse can also provide a series of CEs or a sequence of actions \cite{FACE,OrdCE} that incrementally change users' features to ultimately achieve the desired outcome. Some key desiderata to evaluate CEs are \cite{Review2}: (1) \textit{validity}: whether it gives the desired outcome, (2) \textit{proximity}: how much the changes are measured by a distance function, (3) \textit{sparsity}: how many features are changed, and (4) \textit{realism}: how realistic recourse recommendations are for an individual, including the feasibility of actions.
However, using CEs to find recourse policies does not necessarily result in a sufficiently \textit{\textbf{safe}} recourse policy, because they might ignore the risk of taking actions, which may (probabilistically) leave a person in a worse situation. Such a recourse policy may even be dangerous to suggest. For instance, asking a person to change jobs may result in them losing their current job and being jobless (as illustrated in Recourse Policy B in Figure \ref{fig:stable_plan}). Finding alternatives with lower risks but a slightly higher expected cost may be preferred by an individual. In the context of CE methods, this means that sometimes a more ``distant'' state (set of feature values) may be a better recourse target if the actions required to reach it carry less risk of higher costs.

To explicitly incorporate risk into algorithmic recourse, our work introduces the problem of computing \emph{safe algorithmic recourse} (\textbf{SafeAR}). This has hitherto not been discussed in the literature on algorithmic recourse. The objectives of SafeAR are to suggest different recourse policies with different risk profiles and to empower the affected individual with risk-averse alternatives to decide for themselves.\footnote{SafeAR does not advocate for the policy with the lowest risk as it may have a higher average cost. The emphasis is to provide multiple recourse policies for individuals to choose from, some of which can be safer and more suitable based on their risk tolerance.}
Reinforcement learning (RL) methods can be used to compute such recourse policies. Typically, a policy in RL finds the best action given a (feature) state, which maximizes the expected reward (or minimizes the cost) and can incorporate uncertainty in cost and action effects. To account for risk, we incorporate the variance in costs during policy computation and connect risk-sensitive reinforcement learning ideas ~\cite{Risk_MDP,MDP_cvar,risk_MDP_2} with algorithmic recourse. 
Our contributions are:
\begin{itemize}
    \item Develop the concept of SafeAR by highlighting the value of considering risks in algorithmic recourse, which existing recourse measures do not cover.
    
    \item Formulate algorithmic recourse problems as Finite Horizon Markov Decision Processes (MDPs) and propose a method (Greedy Risk-Sensitive Value Iteration, G-RSVI) to compute risk-aware policies for finite horizon MDPs

    \item Incorporate succinct measures of risk from financial literature to the assessment of algorithmic recourse; these measures are Value at Risk \cite{VaR} and Conditional Value at Risk~\cite{CVaR}. 

    \item Evaluate the policies with different risk profiles computed by G-RSVI on two real-world datasets (UCI Adult Income, German Credit), considering risk measures, sparsity, and proximity; demonstrate that the latter do not implicitly factor in risks. 

    \item Initiate an investigation into gender disparities in terms of risk exposure in the aforementioned datasets.
\end{itemize}

\section*{Motivating Example}
\begin{figure}[!t]
    \centering
    \includegraphics[width=0.46\textwidth]{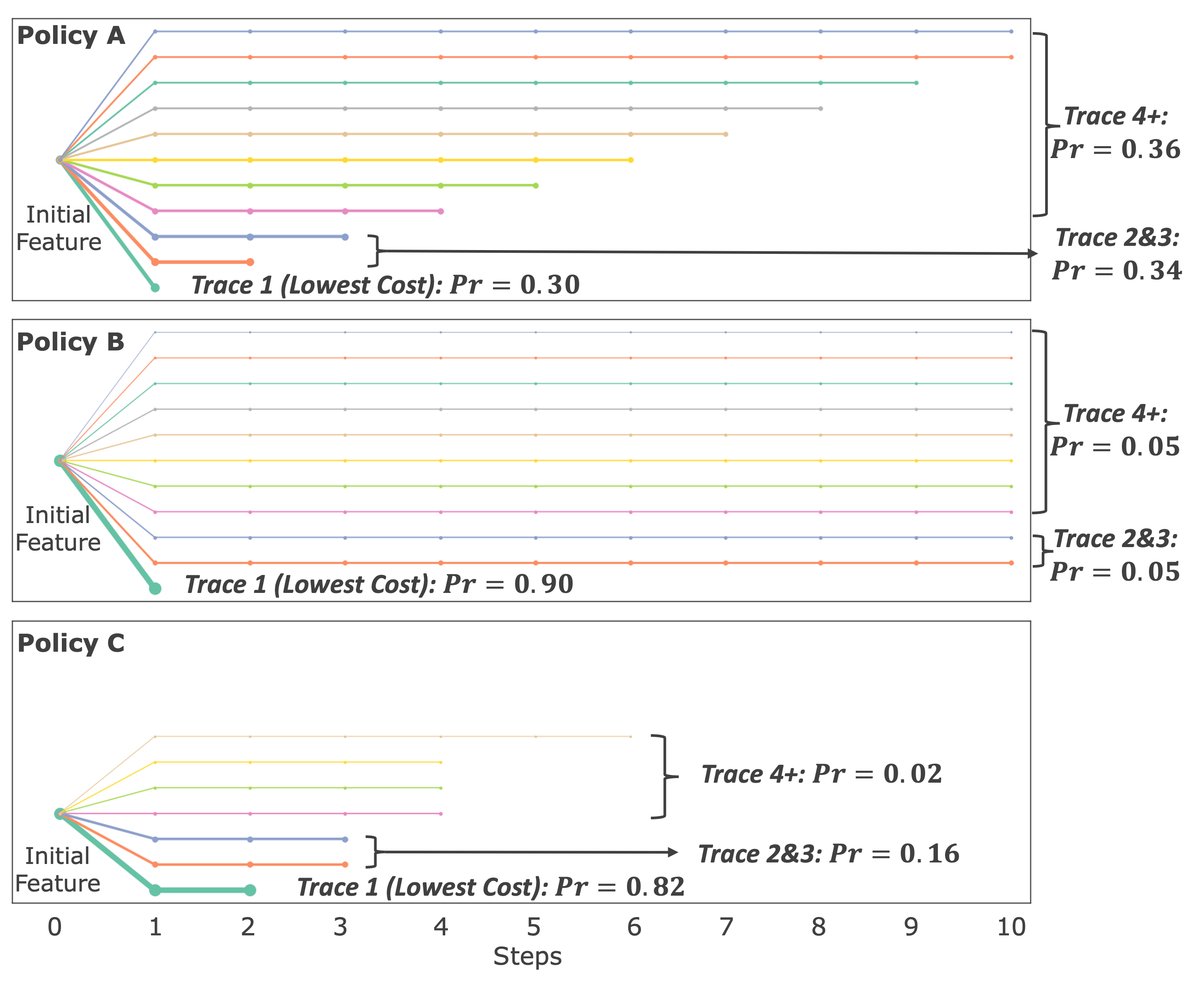}
    \caption{Visualizing risks in three recourse policies from Figure \ref{fig:stable_plan}. The x-axis indicates the cost, and the line (outcome) thickness indicates the outcome probability.}
    \label{fig:stable_plan_viz}
\end{figure}

To better illustrate the concept of SafeAR with risk-aware policies, consider the following motivating example. A company uses a trained black box ML model to determine loan approvals. The model uses a set of features of the loan applicant (housing, job, savings, age, and education) and initially rejects the applicant. In this recourse scenario, the action costs are in terms of time units (months), each action taken has a probability of success, and failure could transition into a less favorable state. Let us look at three recourse policies that could be given to the applicant illustrated in Figure \ref{fig:stable_plan}:

\begin{itemize}
    \item \textit{Policy A: Nearest CE, Expected Cost 3.3}. This could be found by a recourse algorithm that optimizes for feature sparsity. It would require the applicant to \textit{Own-a-House} (one feature change). This policy ignores the uncertainty in the applicant's ability to purchase a house within 1 month (time cost), and there is a $70\%$ chance that the applicant would remain in the same state. Therefore, the expected time cost would be much more than 1 month.
    
    \item \textit{Policy B: Risk-Neutral, Expected Cost 1.5}. This policy recommends the applicant to \textit{Change-Job}, and doing so helps increase the savings and reach the desired outcome with $90\%$ probability. However, there is a small chance ($10\%$) that this action would result in losing their current job and becoming unemployed, from which the cost to recover would be much higher. The expected total cost when considering probabilities is the lowest among the three policies. If optimizing for expected cost alone, this policy would be found and might lead the applicant to a worse situation in which they require a high cost to recover from. This is the type of risk in a recourse policy that a user might want to know about and manage. 
    
    \item \textit{Policy C: Risk-Averse, Expected Cost 2.2}. This policy provides a safer policy to the applicant, where failures do not lead to a worse situation. The actions for this recourse are \textit{Improve-Education-in-Part-Time} and then \textit{Increase-Savings}. The risk associated with this policy is lower than Policy B, but it has a higher expected cost. Policy C might not be found by methods that minimize proximity, as improving educational background could be considered a larger change than getting a higher-paying job. 
\end{itemize}

With the presence of uncertainties, different recourse trajectories (outcomes) might be encountered when following the recourse policy. Figure \ref{fig:stable_plan_viz} visualizes the probabilities of possible outcomes and their associated costs for this example. With the risk-averse Policy C, the applicant is able to receive the desired outcome in 3 time-steps (cost) with probability $98\%$, and the risk of it taking more than 3 is much less than Policy A or B, even if the expected cost is higher than Policy B. Computing such diverse policies in terms of risk and surfacing the risk information to empower the affected individual is the motivation behind SafeAR.

\section*{Related Work}
Existing algorithmic recourse methods \cite{Review4} can be grouped into three categories: one involves finding the nearest CEs as the smallest changes to the individual's feature vector. Solutions focus on \emph{proximity} \cite{wachter2017counterfactual}, \emph{sparsity}, and \emph{diversity} \cite{MACE,CEGuidedByPrototype,DiCE,DACE} using multi-objective optimization \cite{MOCE} and decision trees \cite{CETree}. Also, generative algorithms \cite{REVISE,PlausCF} are used to ensure \emph{plausibility}, by generating CEs within data distributions. These methods do not give a sequence of actions or policy to follow and have no mention of risk.

In the second category, recourse is achieved by recommending a sequence of actions \cite{ARinLinear} or by providing a path over the feature-space along dense regions of the data manifold \cite{FACE} considering \emph{feasibility} and \emph{actionability}. Methods also incorporate \emph{causality} through structural causal models (SCMs) \cite{DirectLiNGAM} to explicitly model inter-variable causal relationships \cite{CausalConstraints} and provide an ordered sequence of CEs \cite{OrdCE,ConsequenceCE}. Lastly, robust recourse methods \cite{ROAR,DiRRA} address the issues for data changes and model parameter shifts. None of these methods consider the risk of higher costs due to the probability of adverse outcomes.

For computing risk-aware recourse policies, we turn to the reinforcement learning (RL) literature. RL methods can provide recourse policies that consider uncertainties in transitions when taking recourse actions. There is existing work that models the recourse problem as MDPs \cite{directiveExplain}. ``ReLAX'' \cite{Relax} generates recourse plans by deep reinforcement learning but under deterministic feature transitions, ignoring uncertainties and thus risk. FASTAR \cite{AmortizedRL} presents a framework that translates an algorithmic recourse problem into a discounted MDP and demonstrates comparable recourse performance as CE methods. Although FASTAR models uncertainties,  it only optimizes for the expected cost and does not incorporate any risk measures. Our first method for SafeAR (G-RSVI) computes recourse policies by considering both the expected cost and the risk of higher costs. 

One way of measuring risks in cost in RL is through the variance in the total cost (over all steps) \cite{sobel1994mean_var_MDP,prashanth2016variance_MDP}. There are also other RL methods that factor in risks in policies~\cite{RSRL,Risk_MDP,borkar2010learning_risk_MDP,chow2015risk_MDP}. To communicate the idea of SafeAR in this work, we use a modification of value iteration (G-RSVI) to incorporate the cost-variance trade-off into the computation to get risk-averse policies. MDPs can naturally incorporate action costs, probabilistic action dynamics, action feasibility, and causal constraints. These can all be personalized to the recipient, as properties like action costs can be unique to each person. To the best of our knowledge, SafeAR is the first attempt to connect the literature on \textit{risk-sensitive} RL to algorithmic recourse.

\section*{Safe Algorithmic Recourse}
\label{sec:SafeAR}
\subsection*{Algorithmic Recourse Problem Statement}
\label{sec:MDP}
Let $f:\mathcal{X}\rightarrow \mathcal{Y}$ be a decision function operationalized by a ML algorithm or model, where $x\in\mathcal{X} = \mathcal{X}_1 \times\dots\times \mathcal{X}_D$ is the set of instances described by $D$ features of an individual, and $\mathcal{Y}=\{y^-, y^+\}$ are the unfavorable and favorable decision outcomes, respectively. An individual with features $x_o$ initially gets an unfavorable outcome $f(x_o)=y^-$, and the general objective of algorithmic recourse is to find actions resulting in a path $x_o,\dots, x^*$ that leads to final feature instance $x^*$ so that $f(x^*)=y^+$. Our work is agnostic to the type of ML model $f$ and only requires the model outputs to be categorized into unfavorable outcomes $y^-$ and favorable outcomes $y^+$. For simplicity of discourse, we use a binary classifier for $f$.

\subsection*{Risk-Aware Recourse Policies Using Finite Horizon Markov Decision Processes}
\label{sec:risk-aware MDP}

To compute SafeAR recommendations, we frame the problem as solving a finite horizon Markov Decision Process (MDP), defined as a tuple of $\left<S, A, T, R, H \right>$.

\subsubsection{States ($S$)}
$S$ is a set of all possible states for individuals in the recourse. Each of the states maps to one instance ($x \in \mathcal{X}$) in the combined feature space (input space) of the decision model $f$. For a valid state space,  there must exist a mapping $g:=S \rightarrow X$, where $\forall s \in S, \exists x\in X$ such that $g(s)=x$.
In this work, we keep the mapping $g$ as one-to-one, meaning the state space is equivalent to the feature space. However, the state space $S$ can be \textit{richer} than the feature space $\mathcal{X}$ because the states and actions for recourse can involve more or different features than the ones used in the decision model $f$. 
For example, ``resting heart rate'' can be a feature in a health insurance premiums calculator $f$, but ``average calories burned'' is not. However, the latter may be in the recourse state, as it is directly affected by actions (e.g., \textit{Exercise}), and in turn can have a causal effect on ``resting heart rate''. Using only the same features as in $f$ may not be adequate for computing recourse policies, as they may not cover the states and actions that a person actually has to change during the recourse. This gives us a reason to expect separate action, transition, and cost models for recourse, rather than assuming they can be extracted from the data used in $f$. Such formulation also enables the recourse to provide personalized recommendations to individuals, as advocated for in ~\cite{venkatasubramanian2020philosophical}.

\subsubsection{Actions ($A$) and Transitions ($T$)}
For the state space $S$, we have a set of feasible actions $a\in A$. The effect of an action can change multiple features. The features can be categorized into three types \cite{Review1}: 1) immutable features (e.g., birthplace), 2) mutable and actionable features (e.g., occupation, bank balance) that define the action space of the recourse, 3) mutable but non-actionable features (e.g., credit score) that cannot be directly modified by an individual. Mutable features can be modified as a consequence of changing other features.
The state transition model would need to capture causal relationships between features and ensure the realism of recourse. The state transition model $T:=p(s'|s,a)$ is defined as the transition probability between two states ($\{s,s'\} \in S\times S)$ given the action $a$. 

\subsubsection{Rewards ($R$)}
$R:=r(s,a,s';f)$ is the reward or cost incurred by reaching state $s'$ by performing an action $a$ at a state $s$. ``Reward'' and $r(.)$ are the typical terms and notations used in RL literature, but rewards can be positive or negative (cost). We will henceforth use ``cost'' in this work since we are focusing on the recourse cost to the recipient, i.e. $r(s,a,s';f)$ tells us the cost incurred to the recipient when the transition $(s,a,s')$ occurs during the recourse. Additionally, when the ML model $f$ gives the favorable outcome in a state ($f(s)=y^+$), then no more actions are needed in the recourse. To capture this, we add a zero-cost action in all favorable (goal) states, transitioning to the same state.

As for the real-world semantics of the cost, it can be a combined measure of multiple factors such as elapsed time, material expenses, opportunity cost, etc. The cost may be averaged across a group or tailored for each person, which requires domain knowledge---so do the feasible actions and the transitions. CE methods such as DiCE \cite{DiCE} and FACE \cite{FACE} also require domain knowledge to design distance (cost) functions, where the function $r(.)$ can be defined in terms of how much the state changes by an action using \emph{sparsity} of feature changes and \emph{proximity} of the recipient's state changes over pre-defined distance functions.

\subsubsection{Horizon ($H$) and Recourse Policies}
Horizon $H$ is the maximum number of steps in the finite horizon MDP, and $h:=[1:H]$ is the step number over the horizon.
A recourse policy is the same as an MDP policy, expressing how to act in each state of each step in the horizon to get to a favorable state. This is formalized as $\pi=(\pi_1\dots\pi_H)$, where $\pi_i:=S\rightarrow A$ maps each state to an action for each step $i$. 

\section*{SafeAR Methodology}
We present a method to compute risk-averse recourse policies and measures to evaluate the risk for SafeAR.

\subsection*{Greedy Risk-Sensitive Value Iteration}
We present a greedy algorithm to compute risk-averse policies by incorporating cost variance. We first define $\hat{R}^\pi_h(s)=\sum_{i=h}^H r(s_i,a_i,s_{i+1})|(s_h=s),\pi$ as the total recourse cost accrued over the horizon $H$ from a rollout obtained by following a policy $\pi$ starting at state $s$ and step $h$. Since outcomes of actions are probabilistic, the total recourse cost is a distribution. In risk-neural settings, the recourse policy $\pi$ maximizes the expected total cost $\mathbb{E}[\hat{R}^\pi_h(s)]$ or the mean value $\mu[\hat{R}^\pi_h(s)]$. G-RSVI also considers the variance in the total recourse cost to manage risk and seeks to find a policy $\pi$ to maximize the following value function for each state $s$ starting at the first step $h=1$:
\begin{equation}
\label{eq:mean-variance value}
    V^\pi_1(s)=\mu(\hat{R}^\pi_1(s))-\beta \cdot \sigma(\hat{R}^\pi_1(s)).
\end{equation}
We denote $V^\pi_h(s)$ as the risk-sensitive value of state $s$ in step $h$ by following policy $\pi$. $\beta\geq 0$ is the tuning parameter that represents the risk profile, and a higher value means more risk averse. When $\beta=0$, the problem reduces to finding the policy with the least expected cost only, which is the standard optimization objective in MDPs. Here, $\sigma$ returns the standard deviation of the total cost, and $\sigma^2$ returns the variance.
In G-RSVI, we optimize Equation \ref{eq:mean-variance value} by greedily maximizing $V^\pi_h(s)$ at each step starting from the end step $H$ and moving backward to the first step. At each step $h$, the action is selected to maximize the risk-sensitive value using: 
\begin{equation}
\label{eq:greedy_value_iteration}
    V_h=\underset{a}{\max}\ \mu[r(\cdot)+V_{h+1}]-\beta\sigma[r(\cdot)+V_{h+1}],
\end{equation}
where $V_{h+1}$ is the value computed in the previous step. The risk-sensitive action value or Q-value $Q_h(s,a)$ at step $h$ is defined as:
\begin{align}
\label{eq:mean-variance Q-value}
    Q_h(s,a) & =\mathbb{E}_{s'}[r(s,a,s')+V_{h+1}(s')] \nonumber\\
    & -\beta\sigma[r(s,a,s')+V_{h+1}(s')].
\end{align}

\begin{algorithm}[t]
\caption{G-RSVI}
\hangindent=9mm \textbf{Input:} recourse MDP$\langle S,A,T,R,H \rangle$, ML model $f$ 

\hangindent=17.6mm \textbf{Parameters:} risk aversion level $\beta\in[0,\infty]$

\begin{algorithmic}[1]
\label{alg:RAVI}
\STATE $V_{H+1}(s)\leftarrow 0, \forall s \in S$
\FOR{step $h=H,H-1,\dots,1$}
\FOR{each state $s \in S$}
\FOR{each action $a \in A$}
\STATE $r(s')\leftarrow$ get reward $R(s,a,s';f)$
\STATE $p(s')\leftarrow$ get transition probability $T(s'|s,a)$
\STATE $\mu\leftarrow\sum_{s'}p(s')[r(s')+V_{h+1}(s')]$
\STATE $\sigma^2\leftarrow\sum_{s'}p(s')[r(s')+V_{h+1}(s')-\mu]^2$
\STATE $Q_{h}(s,a)\leftarrow \mu-\beta\sigma$ (Equation \ref{eq:mean-variance Q-value})
\ENDFOR
\STATE $V_{h}(s)\leftarrow \max Q_{h}(s,\cdot)$
\STATE $\pi_h(s) \leftarrow \argmax Q_h(s,\cdot)$
\ENDFOR
\ENDFOR
\RETURN recourse policy $\pi_h(s)$
\end{algorithmic}
\label{alg:G-RSVI}
\end{algorithm}

If only optimizing the expected reward, this procedure would find the optimal policy because the optimal sub-structure assumption for dynamic programming holds. However, G-RSVI does not guarantee to find the optimal policy for Equation \ref{eq:mean-variance value}. It does, however, provide one straightforward way to incorporate risks into recourse policy computation, and it completes computation with a single sweep over the state and horizon space. There are a variety of heuristic methods one can use with different trade-offs to compute risk-aware policies. In this work, to focus on the exposition of the concept of SafeAR, we limit our approach to discrete states, discrete actions, and finite horizon MDPs.

Our G-RSVI algorithm is shown in Algorithm \ref{alg:G-RSVI}. We compute the policy by sweeping backward from the last horizon step (Line 2). For all state-action pairs at each step, the action values $Q_h(s,a)$ are computed by Equation \ref{eq:mean-variance Q-value} (Lines 5-9). The best action for each state in each step is then chosen by the one with maximal $Q_h(s,a)$. It also gives us the state value $V_h(s)$ and the policy for each step (lines 11, 12). Other ways of scoring values and selecting actions can be used in our algorithm. For example, one can also penalize Lower Partial Standard Deviation (LPSD)  \cite{LPSD} (Appendix A.8) of lower-than-average values (negative) to avoid higher-than-average costs, as greater negative values indicate higher costs.
Equation \ref{eq:mean-variance Q-value} helps us compare against risk-neutral policies that optimize for expected value only, by setting $\beta=0$. We leave the analysis of different risk-sensitive algorithms for recourse to future work.

\subsection*{Risk Measures for Recourse Policies}
To evaluate the risk associated with a recourse policy, we propose the following measures. 

\subsubsection{Success Rate $\pmb{(\rho_H)}$:}
It estimates the probability of success within the horizon $H$ by following the recourse policy. For example, $\rho_5=0.9$ means a favorable outcome state will be reached within $5$ steps 90\% of the time. $\rho_H$ is not equivalent to \emph{validity}, which only determines whether a feature instance with a favorable decision exists. $\rho_H$ is affected by the uncertainty of action outcomes in the recourse policy.

\subsubsection{Mean-Variance Cost $\pmb{(\mu_{cost},\sigma^2_{cost})}$:}
It computes the expected value and variance in the total cost of following recourse policies. Since the distribution of costs is not necessarily Gaussian, these statistics can be misleading or hard to interpret. Hence, we propose additional measures.

\subsubsection{Value at Risk (VaR$\pmb{_\alpha}$):}
VaR \cite{VaR} is to provide a succinct probabilistic guarantee on the recourse policy cost. We evaluate VaR \cite{VaR} of the recourse cost to answer the question ``What is the highest cost at a given level of cumulative probability (confidence level)''.  For example, VaR$_{95}= 5.6$ means that with $95\%$ probability, the recourse cost is at most $5.6$.
Formally, assuming the total cost of recourse $x_c$ is the value of a random variable $X_c$ with a cumulative probability distribution $F_{X}(x_c)$, under confidence level $\alpha\in[0,1]$ VaR$_\alpha$ is computed as:
\begin{equation}
    \text{VaR}_\alpha(X_c)=\min\{x_c|F_X(x_c)\geq \alpha\}.
\end{equation}

\subsubsection{Conditional Value at Risk (CVaR$\pmb{_\alpha}$):}
CVar~\cite{CVaR} is a complementary measure to VaR and tells us the expected worst-case cost when the cost exceeds the threshold given by VaR$_\alpha$ value. For example, CVaR$_{95}=8.4$ means that when the cost exceeds the 95-percentile cost, the average cost for those cases is $8.4$. It is computed as:
\begin{equation}
    \text{CVaR}_{\alpha}=\mathbb{E}[x_c|x_c>\text{VaR}_{\alpha}].
\end{equation}

\section*{Experimental Results}
\label{sec:experiments}

\begin{table*}[!t]
    \fontsize{10}{11}\selectfont
    \centering
    \begin{tabular}{c|l|c|c|cc|cc|c|c}
         \hline
         \multicolumn{1}{c}{\textbf{Dataset}} &
         \multicolumn{1}{c}{\textbf{Policy}} & \multicolumn{1}{c}{$\pmb{\rho_{H=12}}$}& \multicolumn{1}{c}{$\pmb{(\mu_{cost},\sigma^2_{cost})}$}
         &
         \multicolumn{1}{c}{\textbf{VaR}{$\pmb{_{80}}$}}&
         \multicolumn{1}{c}{\textbf{CVaR}{$\pmb{_{80}}$}}&
         \multicolumn{1}{c}{\textbf{VaR}{$\pmb{_{95}}$}}&
         \multicolumn{1}{c}{\textbf{CVaR}{$\pmb{_{95}}$}}&
         \multicolumn{1}{c}{\textbf{Spars.}}&
         \multicolumn{1}{c}{\textbf{Proxi.}}\\
         \hline
        \multirow{5}{2.8cm}{\makecell{\textbf{Adult Income}\\($n=25923$)\\~\\\vspace{-3mm}\\~\\(Example Instance)}}
        &$\beta=0$    & 0.994 & (\textbf{3.49}, 1.23)   & 3.81           & 6.31           & 4.76          & 7.53          & \textbf{2.09} & \textbf{2.87}\\
        &$\beta=0.25$ & 0.994 & (3.51, 0.89)            & \textbf{3.64}  & \textbf{6.10}  & 4.46          & 7.43          & 2.16          & 3.06\\
        &$\beta=0.5$  & 0.993 & (3.59, \textbf{0.77})   & 3.66           & 6.11           & \textbf{4.44} & \textbf{7.40} & 2.21          & 3.18\\
        \cline{2-10}
        &$\beta=0$    & 1.000 & (\textbf{4.63}, 1.86)   & 5.80           & 8.54           & 6.80          & 9.80          & 3.92          & \textbf{3.92}\\
        &$\beta=0.5$  & 1.000 & (4.79, \textbf{0.13})   & \textbf{4.80}  & \textbf{6.80}  & \textbf{4.80} & \textbf{6.80} & \textbf{3.00} & 4.86\\
        &$\beta=0.75$  & 1.000 & (4.79, \textbf{0.13})   & \textbf{4.80}  & \textbf{6.80}  & \textbf{4.80} & \textbf{6.80} & \textbf{3.00} & 4.86\\

        \hline
        \hline
        \multirow{6}{2.8cm}{\makecell{\textbf{German Credit}\\($n=281$)\\~\\~\\\vspace{-3mm}\\(Example Instance)}}         
        &$\beta=0$    & 1.000 & (\textbf{1.65}, 0.48)   & 1.96           & 3.66           & 2.63          & 4.56          & \textbf{1.26} & \textbf{1.33}\\
        &$\beta=0.25$  & 1.000 & (1.67, 0.35)            & \textbf{1.87}  & \textbf{3.51}  & 2.51          & 4.50          & 1.34          & 1.43\\
        &$\beta=0.5$  & 1.000 & (1.70, \textbf{0.30})   & 1.90           & 3.56           & \textbf{2.48} & \textbf{4.40} & 1.40          & 1.50\\
        \cline{2-10}
        &$\beta=0$    & 1.000 & (\textbf{2.48}, 3.49)   & \textbf{4.00}  & 6.13           & 7.00          & 8.33          & \textbf{1.00} & \textbf{1.00}\\
        &$\beta=0.5$  & 1.000 & (2.81, 1.19)            & \textbf{4.00}  & \textbf{5.44}  & \textbf{5.00} & \textbf{6.00} & \textbf{1.00} & 2.00\\
        &$\beta=0.75$ & 1.000 & (3.87, \textbf{0.65})   & 4.40           & 5.50           & 5.40          & 6.67          & 2.00          & 3.00\\
         \hline
    \end{tabular}
    \caption{Evaluating recourse policies with different risk-aversion levels $\beta$ and horizon $H=12$ for AID, GCD, and two example instances sampled from the datasets. $n$ denotes the number of instances used for evaluation. For each risk-aversion level, we compute a recourse policy and apply it to each instance (initially with the undesired outcome) in the dataset. Due to the uncertainties in action outcomes, we estimate the average of each measure using simulated 100 trials. Then, the average across all instances for each metric is computed. The best metric values among the policies are highlighted in bold.}
\label{tb:dataset_risk_measures}
\end{table*}

Motivated by the datasets used in the algorithmic recourse literature, we evaluate our method on the following two datasets: Adult Income Dataset (AID) (32561 data points) \cite{uci_adult} and German Credit Dataset (GCD) (1000 data points) \cite{uci_credit} and show how risk measures vary with different recourse policies. In AID, the recourse is to help individuals earn an income greater than 50,000. In GCD, the recourse is to help get a loan approval by reaching a good credit standing. Here we consider the version of GCD \cite{credit_small} with 9 features. To process the datasets for G-RSVI, we convert continuous feature values into discrete values (details included in Appendix A.1).
We then train random forest classifiers for both datasets. Dataset features, feature state dimensions, and classifier accuracies are reported in Table \ref{tb:dataset}. G-RSVI does not access the dataset when computing the policy and only uses the trained ML model to indicate whether the desired outcome is received.\footnote{\label{ftn:suppl_location}All supplemental materials (appendices and code implementations) are available through \url{arxiv.org/abs/2308.12367}.}


\begin{table}[htp]
    \fontsize{10}{11}\selectfont
    \centering
    \begin{tabular}{c|c|c|c}
         \hline
         \multicolumn{1}{c}{\textbf{Dataset}} &
         \multicolumn{1}{c}{\textbf{Immutable}} &
         \multicolumn{1}{c}{\textbf{\#States}}&
         \multicolumn{1}{c}{\textbf{ML Model}}\\
         \multicolumn{1}{c}{\textbf{(\#Feat.)}} &
         \multicolumn{1}{c}{\textbf{Features}} &
         \multicolumn{1}{c}{\textbf{ }}&
         \multicolumn{1}{c}{\textbf{(Accuracy)}}\\
         \hline
         AID & Gender, Race, & 57600 & Rand.Forest \\
         (8) & Marital Status & & (0.81) \\
         \hline
         GCD & Gender, Purpose, & 147456 & Rand.Forest \\
         (9) & Credit Amount & & (0.76) \\
         \hline
    \end{tabular}
    \caption{Dataset Overview}
    \label{tb:dataset}
\end{table}


\begin{table*}[ht]
    \fontsize{10}{11}\selectfont
    \centering
    \begin{tabular}{c|l|c|c|cc|cc|c|c}
         \hline
         \multicolumn{1}{c}{\textbf{Dataset}} &
         \multicolumn{1}{c}{\textbf{Policy}} & \multicolumn{1}{c|}{$\pmb{\rho_{H=12}}$}& \multicolumn{1}{c|}{$\pmb{(\mu_{cost},\sigma^2_{cost})}$}
         &
         \multicolumn{1}{c}{\textbf{VaR}{$\pmb{_{80}}$}}&
         \multicolumn{1}{c}{\textbf{CVaR}{$\pmb{_{80}}$}}&
         \multicolumn{1}{c}{\textbf{VaR}{$\pmb{_{95}}$}}&
         \multicolumn{1}{c}{\textbf{CVaR}{$\pmb{_{95}}$}}&
         \multicolumn{1}{c}{\textbf{Spars.}}&
         \multicolumn{1}{c}{\textbf{Proxi.}}\\
         \hline
        \multirow{6}{*}{\makecell{\textbf{Adult Income}\\(Female, $n=9824$)\\~\\~\\\vspace{-3mm}\\(Male, $n=16099$)}}
        &$\beta=0$    & 0.988 & (\textbf{4.56}, 1.41)   & 4.76           & 7.07           & 5.70          & 8.00          & 2.58          & \textbf{3.77}\\
        &$\beta=0.25$  & 0.987 & (4.57, 1.18)            & \textbf{4.64}  & \textbf{6.95}  & \textbf{5.48} & \textbf{7.84} & \textbf{2.55} & 3.87\\
        &$\beta=0.5$  & 0.985 & (4.61, \textbf{1.11})   & 4.66           & 6.99           & 5.51          & 7.89          & \textbf{2.55} & 3.93\\
        \hhline{|~|---------|}
        &$\beta=0$    & 0.997 & {(\textbf{2.84}, 1.12)}  & 3.27           & 5.79           & 4.30          & 7.26          & \textbf{1.79} & \textbf{2.32}\\
        &$\beta=0.25$  & 0.997 & (2.87, 0.72)            & {\textbf{3.08}}  & 5.51           & 3.95          & 7.15          & 1.93          & 2.56\\
        &$\beta=0.5$  & 0.998 & {(2.98, \textbf{0.57})}   & 3.10           & {\textbf{5.48}}  & {\textbf{3.92}} & {\textbf{7.08}} & 2.01          & 2.73\\
        \hline
        \hline
        \multirow{6}{*}{\makecell{\textbf{German Credit}\\(Female, $n=103$)\\~\\~\\\vspace{-3mm}\\(Male, $n=178$)}}
        &$\beta=0$    & 1.000 & (\textbf{1.72}, 0.56)   & 2.08           & 3.70           & 2.84          & 4.66          & \textbf{1.25} & \textbf{1.36}\\
        &$\beta=0.25$  & 1.000 & (1.75, 0.38)            & \textbf{2.00}  & \textbf{3.61}  & 2.67          & 4.63          & 1.33          & 1.47\\
        &$\beta=0.5$  & 1.000 & (1.79, \textbf{0.33})   & 2.02           & 3.64           & \textbf{2.59} & \textbf{4.51} & 1.41          & 1.57\\
        \hhline{|~|---------|}
        &$\beta=0$    & 1.000 & (\textbf{1.61}, 0.43)   & 1.89           & 3.64           & 2.61          & 4.62          & \textbf{1.27} & \textbf{1.32}\\
        &$\beta=0.25$  & 1.000 & (1.62, 0.37)            & {\textbf{1.81}}  & {\textbf{3.50}}  & 2.41          & 4.42          & 1.35          & 1.41\\
        &$\beta=0.5$  & 1.000 & {(1.65, \textbf{0.29})}   & 1.84           & {\textbf{3.50}}  & {\textbf{2.33}} & {\textbf{4.29}} & 1.40          & 1.46\\
         \hline
    \end{tabular}
    \caption{Evaluating recourse policies across gender for Adult Income and German Credit datasets; the same evaluation procedures are followed as Table \ref{tb:dataset_risk_measures}. The male group is exposed to less risk under the policy with the same risk-aversion level.}
\label{tb:dataset_risk_measures_gender}
\end{table*}

\subsubsection{Transitions and Rewards}
We use qualitative assumptions (domain knowledge) on relative differences in action costs and success likelihood to define the action costs $r(\cdot)$ and transition model $p(\cdot)$. Similar to FASTAR \cite{AmortizedRL}, we assume \textit{improve-education} or \textit{improve-skill} actions would lead to an age increase as causal constraints, and we treat ``Age'' as a mutable but non-actionable feature. These two actions require more time and effort, and therefore the action cost would be larger than other actions such as to \textit{increase-work-hours}. The transition probabilities are heuristically set by domain knowledge. For example, the probability of earning a Ph.D. degree is lower than earning a Bachelor's. For more details on the model values, we refer the reader to Appendix A.2
for an exhaustive list of model transition probabilities and costs. Results from a different model using the same qualitative assumptions are also provided in Appendix A.5
to show the G-RSVI method and results are not specific to a single model.

\subsubsection{Baselines}
To our knowledge, our work is the first to address risks in algorithmic recourse. Among the existing recourse approaches, CE methods do not naturally allow probabilities into the formulation, and FASTAR \cite{AmortizedRL} formulates recourse problems as MDPs \textit{and} allows for stochastic transitions. FASTAR sets rewards in terms of distance measures between states. No matter what reward function is used---either distance-based or user-defined cost---and how transition probabilities are defined---either extracted from a dataset or tuned domain knowledge---FASTAR only seeks to find the recourse policy that maximizes the expected total rewards (risk-neural). This is what a standard algorithm for MDP (value or policy iteration) would find. In our experiments, the policy that maximizes expected total reward corresponds to the risk-neutral policy ($\beta = 0$), and this is the baseline which risk-averse policies compare against. We select $\beta=0.25, 0.50, 0.75$ for generating risk-averse recourse policies, and higher $\beta$ indicates higher risk-aversion.

\subsubsection{Performance Evaluation}
Table \ref{tb:dataset_risk_measures} reports the risk measures for each experimental setting. The horizon is set to 12. We also present sparsity ($L_0$ distance) and proximity ($L_0$ distance for nominal features + $L_1$ distance for ordinal and numerical features) between initial and final states. All measures are averaged over the entire dataset, as well as the measures for two example instances (sampled from $\mathcal{X}$). Recourse policies are computed using the same cost and transition functions for all instances in the dataset. In the results, we see that for both datasets, more risk-averse policies (higher $\beta$ values) can provide recourse with less variance in cost $\sigma^2_{cost}$ but often require a higher mean cost $\mu_{cost}$. For the same $\alpha$ confidence level, risk-averse policies give lower costs in VaR and CVaR than risk-neutral policies. Also, for the example instance in GCD, variance in cost $\sigma^2=0.65$ at risk-aversion level $\beta=0.75$ is significantly lower than the $\sigma^2=3.49$ at $\beta=0$. For the example instance in AID, increasing risk-aversion to $\beta=0.75$ would not find a different policy than $\beta=0.5$, which can happen if the same relative ordering of state values $V_h(s)$ is found at each step. In the results, we observe that low sparsity and proximity do not correspond to risk-averse policies, meaning optimizing for them would not necessarily factor in risks.

\subsubsection{Visualizing Risks in Recourse Policies}
\begin{figure}[!t]
    \centering
    \includegraphics[width=0.47\textwidth]{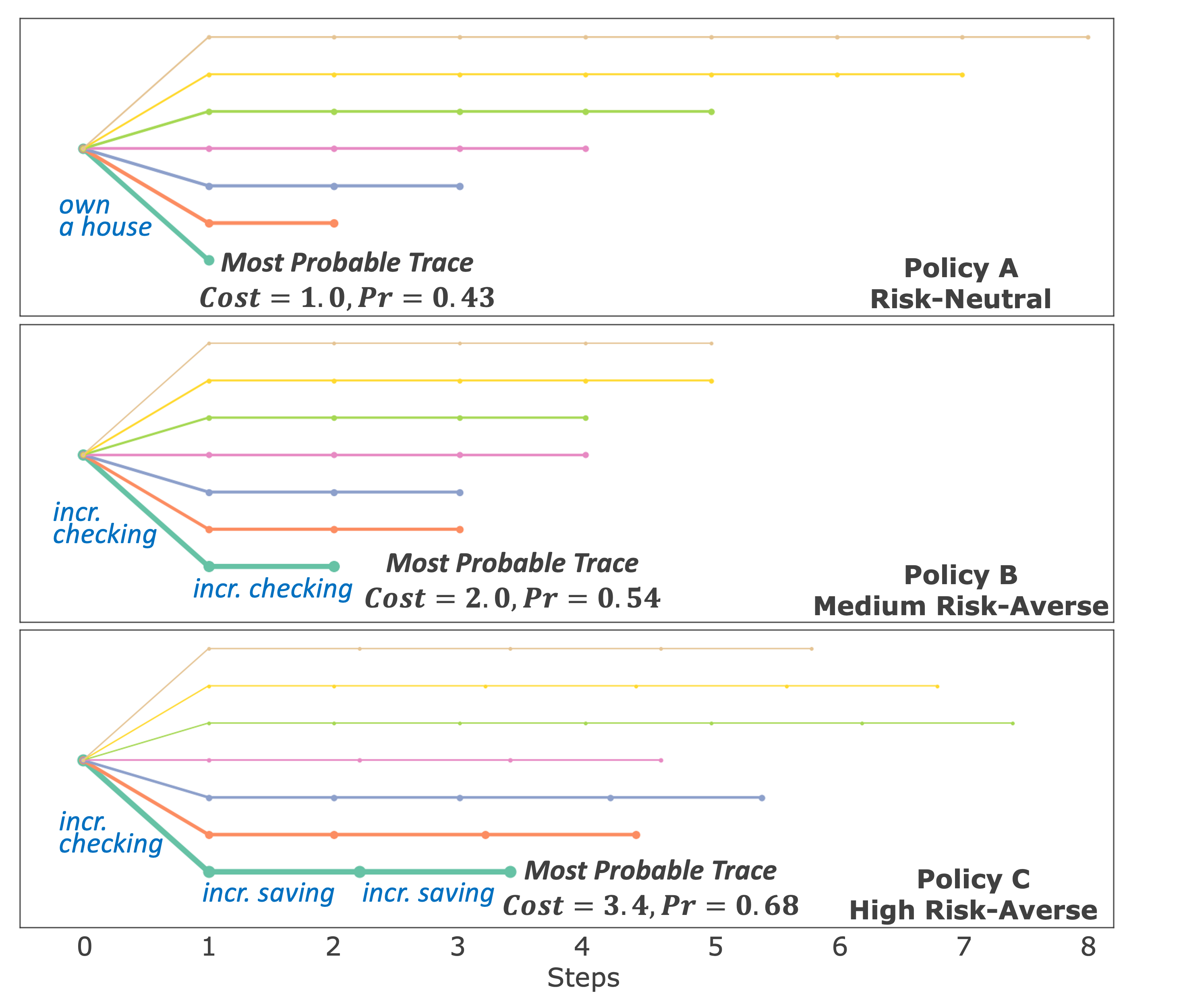}
    \caption{Policy visualization Example in German Credit.}
    \label{fig:credit_policy_viz}
\end{figure}

In Figure \ref{fig:credit_policy_viz}, we use our policy-risk visualization for a set of policies from the GCD dataset. For each policy, we show the most probable outcomes (rollouts), the length of each trace corresponds to the total cost, and the thickness of each trace corresponds to the probability of the outcome. This approach visualizes the variability in cost, which can help a person get an intuition of their risk in addition to the recommended actions. 

\subsubsection{Exploring Risk Disparity}
Inspired by prior work that investigated recourse fairness \cite{CERTIFAI,RAGUEL,FainessCausalAR,raimondi2022equality}, we now look at the disparity that may exist in risk measures across two gender groups (male and female) provided in AID and GCD. Table \ref{tb:dataset_risk_measures_gender} reports the risk measures (averaged) for female and male groups in both datasets. The p-values for statistical significance of the difference between the groups are provided in Appendix A.4.
We define disparity in VaR between the two gender groups by following the same recourse policy computed with a risk-aversion level $\beta$ as:
$
\Delta\text{VaR}_{95}^{\beta}=|\text{VaR}_{95}^{\text{Female}}-\text{VaR}_{95}^{\text{Male}}|.
$ The disparity for other measures is similarly computed. 

All measures are in favor of the male group
for both datasets, meaning that for the policy with the same risk-aversion level given to females and males, we expect females would get higher variance in cost ($\sigma^2_{cost}$), higher cost at the VaR confidence level of $\alpha = 80, 95$, and higher costs in the expected worst case scenarios (CVaR) for those confidence levels.
In AID, when increasing the risk-aversion, the disparity of risk measures between two groups becomes \textit{larger}.
We observe $\Delta\text{VaR}_{95}^{\beta}$ increases from 1.4 to 1.59 as $\beta$ increases, and similar trends are observed for the difference in $\sigma^2_{cost}$ and CVaR in AID. This trend indicates that the more we want to achieve risk-aversion, the greater the disparity in risk exposure between the two gender groups in AID. However, in GCD, the difference in $\sigma^2_{cost}$ and $\Delta\text{VaR}_{80}^{\beta}$ do not consistently increase with increased risk-aversion. The disparity between males and females across all measures of risk still exists in GCD. We recall that the same action costs and transitions are used for both males and females. The only difference is the decisions made by the model $f$ for different groups, which affects the number of steps to reach the favorable state ($f(x)=y^+$). This is something that recourse providers may want to keep in mind, and it motivates further discussion on risk disparity in algorithmic recourse.

\section*{Discussion and Conclusions}
The motivation behind our Safe Algorithmic Recourse (SafeAR) is to offer recourse policies with different risk profiles. This enables affected individuals to be aware of the risks and helps them make an informed decision based on their risk tolerance. We connect ideas from risk-sensitive reinforcement learning with the algorithmic recourse literature and propose an algorithm G-RSVI that can provide risk-averse recourse policies for individuals with different risk profiles. In our experiments with the AID and GCD datasets, we showed that the recourse policies generated by G-RSVI were better in terms of the risk measures as compared to the existing risk-neutral approaches. The policy risk was evaluated through cost-variance, VaR, CVaR, and success rate. In addition, we observed that policies with better sparsity and proximity scores need not correspond to risk-averse policies. Lastly, in our experiments, we observed discrepancies between gender groups in risk measures for the same risk-aversion setting, which motivates further studies on recourse fairness in terms of risk exposure. 


\section*{Ethical Statement}
SafeAR would work best with personalized action costs and action success likelihoods for an individual (no personal data was collected or used in this work). If done that way, it would effectively mean asking for personal information. However, the personal data can be deleted after computing the recourse policy, as SafeAR only requires individual action-model data for computation. Also, the black-box ML model does not need information on personal models and preferences on actions and transitions.

\section*{Acknowledgements}
This paper was prepared for informational purposes by the Artificial Intelligence Research group of JPMorgan Chase \& Co. and its affiliates (``JP Morgan''),
and is not a product of the Research Department of JP Morgan. JP Morgan makes no representation and warranty whatsoever and disclaims all liability, for the completeness, accuracy or reliability of the information contained herein.
This document is not intended as investment research or investment advice, or a recommendation, offer or solicitation for the purchase or sale of any security, financial instrument, financial product or service, or to be used in any way for evaluating the merits of participating in any transaction, and shall not constitute a solicitation under any jurisdiction or to any person, if such solicitation under such jurisdiction or to such person would be unlawful.

\bibliography{ref.bib}

\clearpage

\appendix
\section{Evaluation}
Please contact us for the source code used for the experiments and evaluations.

\subsection{Pre-processing Datasets}
\label{apd:data_preprocess}

To prepare the Adult Income Dataset (AID) \cite{uci_adult}, we use the same features as DiCE \cite{DiCE} and follow the same process as \cite{adult_preprocess} used by DiCE. Income $<$50K is considered as unfavorable, and $>$50K is favorable, indicated as $\{0, 1\}$ respectively. In addition, there are two features with integer values: Age and Hours-per-Week (Hrs/Week), and we categorize them into ordinal values as the requirement for our approach, G-RSVI. Similarly, we categorize Credit-Amount (Credit\#), and Duration in the German Credit Dataset (GCD) \cite{credit_small}. The resultant feature lists and feature categories/levels are reported in Table \ref{tb:dataset_preprocess}.

\begin{table}[!htp]
    \fontsize{9}{12}\selectfont
    \centering
    \begin{tabular}{c|l}
    \hline
        \multicolumn{2}{c}{\textbf{Adult Income Feature Values \hfill Target: Income=\{0,1\}}}\\
         \hline
         Age        & $<$20, $<$25, $<$30, $<$40, $<$50, older\\
         Education  & School, HS, Bachelors, Masters, Doc.\\
         Hrs/Week   & Less, Part-Time, Full-Time, Over \\
         Workclass  & Gov., Self, Private, Other\\
         Occupa.    & Prof., WhiteCol, Service, BlueCol, Sales, Other \\
         \textbf{Marital}    & Single, Married, Divorced, Separated, Widowed \\
         \textbf{Race}       & White, Other \\
         \textbf{Gender}     &  Female, Male \\
         \hline
         \hline
        \multicolumn{2}{c}{\textbf{German Credit Feature Values \hfill Target: Risk=\{0,1\}}}\\
        \hline
        Age         & Student, Young, Adult, Senior \\
        Skill       & Little(Non-Res), Little(Res), Skilled, Manage. \\ 
        Savings     & None, Little, Moderate, Rich \\
        Checkings   & None, Little, Moderate, Rich \\
        Duration    & $<$1yr, $<$2yr, $<$3yr, More \\
        Housing     & Free, Rent, Own \\
         \textbf{Purpose}    & TV, Education, Equip., Car, Business, Other \\
        \textbf{Credit\#} & Low, Med, High, High Plus \\
         \textbf{Gender}     &  Female, Male \\
         \hline
    \end{tabular}
        \caption{The feature list and discrete values for both datasets. Bolded features are immutable. The other features are mutable, and Age is non-actionable.}
\label{tb:dataset_preprocess}
\end{table}

\subsection{Transition and Cost Functions}
\label{apd:trans_cost_detail}
We direct the reader to Table \ref{tb:trans_cost_detail} in which we report the action space, action costs (rewards), and the corresponding transitions in terms of action success likelihood used for AID and GCD evaluation in Section \ref{sec:experiments}. When an action fails, we just assume the user stays in the same state. We assume each action can only change one feature, but Age will increase when taking Impr-Edu (improve education) or Impr-Skill (improve skill). For ordinal features, we only allow the state transition to increase by one level; for example, we do not allow Hrs/Week to jump from ``Part-Time'' to ``Over'' (overtime). The costs and probabilities are set based on qualitative assumptions and basic domain knowledge; this is used to determine which actions have higher costs than others. Ideally, these recourse models (action cost and transitions) should be designed for each individual based on their situation. For the purpose of evaluating and showing different recourse policies with different risk-aversion levels, we use the same cost and transition models for the entire dataset. We show how our method produces risk-aware policies for specific instances in Section \ref{apd:instances}, and another set of models preserving similar qualitative assumptions is evaluated in Section \ref{apd:param_reward_trans}.

\begin{table}[!htp]
    \fontsize{9}{12}\selectfont
    \centering
    \begin{tabular}{l|cc}
    \hline
         \multicolumn{3}{c}{\textbf{Adult Income}}\\
         \hline
         \makecell{Action\\Space} & \makecell{Action Costs\\$r(a)$} & \makecell{Transitions\\(Prob. of Success) $p(s')$}\\
         \hline
         Impr-Edu       & 2.0 & HS:1.0, Bachelors:0.9, \\ & &  Masters:1.0, Doc. 0.8 \\
         Incr-Hrs       & 0.8 & 1.0 \\
         Work-Gov       & 1.1 & 0.7 \\
         Work-Self      & 1.0 & 0.5 \\
         Work-Private   & 1.2 & 0.9 \\
         Job-Prof       & 1.0 & 0.5 \\
         Job-WhiteCol   & 1.0 & 0.5 \\
         \hline
         \hline
         \multicolumn{3}{c}{\textbf{German Credit}}\\
         \hline
        Impr-Skill      & 1.5 & Skilled:0.8, Manage:0.8 \\ 
        Incr-Savings    & 1.2 & 0.95 \\
        Incr-Checkings  & 1.0 & 0.7 \\
        Incr-Duration   & 1.0 & 1.0 \\
        House-Free      & 1.5 & 1.0 \\
        House-Rent      & 1.2 & 0.7 \\
        House-Own       & 1.0 & 0.4 \\
         \hline
    \end{tabular}
        \caption{Action Cost and Transition Models; If the action fails, the state stays the same.}
\label{tb:trans_cost_detail}
\end{table}

\subsection{Incorporating Counterfactual Explanations}
Recall that CE methods find the final feature instances or a path of feature instances to achieve recourse. CE methods do not consider feature change uncertainties and transition probabilities, so our method (G-RSVI) cannot be directly compared with CE methods. However, as an approach to incorporate CE methods, we can use the final feature instances provided by CE methods as goal or objective states in our MDP; these are treated as the only favorable outcome states and then we compute the policies by G-RSVI. Here, we first generate 10 CEs using DiCE \cite{DiCE} (denoted as DiCE10) and apply G-RSVI to compute the risk-neutral policy that leads to DiCE10 goal states with favorable outcomes (DICE10$_{\beta=0}$). The cost and transition functions are still required as they model the costs incurred by taking actions and uncertainties of feature changes for users. The performance evaluation is reported in Table \ref{tb:dataset_risk_measures_100}. We notice that using DICE10$_{\beta=0}$ has the lowest sparsity, but other measures are outperformed by policies that explore the full state space, especially the recourse success rate. These results imply that CE methods could potentially prevent individuals from getting risk-averse recourse policies if we limit the target/goal states based on CE methods. This is not surprising since CE methods are not privy to information on uncertainty; this helps emphasize the need to consider an explicit model of recourse actions for risk-averse recourse, rather than use CE methods for recourse.


\subsection{Disparity in Risks across Gender}
\label{apd:gender_p_values}
We perform \emph{Mann–Whitney U test} \cite{mann-whitney} to see if the disparity in the mean value of risk measures across the two gender groups in AID and GCD datasets is statistically significant. We present the p-values for significance in the for cost-mean $\mu_{cost}$ and cost-variance $\sigma^2_{cost}$ in Table in Table \ref{tb:gender_p_values}. We first measure cost-mean and cost-variance for each instance in the entire dataset and then group the measures into two gender groups for the Mann-Whitney U test.
We do the same tests for VaR$_{80}$ and CVaR$_{80}$ in Table \ref{tb:gender_p_values_var_cvar_80}, and for VaR$_{95}$ and CVaR$_{95}$ in Table \ref{tb:gender_p_values_var_cvar_95}. There is a clear statistically significant difference between gender groups for AID, but not for GCD due to limited data points available in the dataset.

\subsection{Parametric Analysis: Using a Different Recourse Model of Costs $R$ and Transitions $T$}
\label{apd:param_reward_trans}

We conduct an additional experiment on risk measures and disparity in risks across gender groups using a different set of action costs and transitions---these follow the same qualitative assumptions as previously discussed---and report the results in Table \ref{tb:trans_cost_detail_env2}. This set of models presumes similar qualitative relationships as Table \ref{tb:trans_cost_detail} with adjusted cost magnitude and action success likelihoods. The results for this different recourse model are summarized in Table \ref{tb:env2_risk_measures_gender}. From this set of results, G-RSVI can still compute risk-averse policies that have lower variance in cost $\sigma_{cost}^2$, lower VaR, and lower CVaR, compared to risk-neutral policies. Also, we observe similar gender disparity as with the first recourse action model---the risk measures are in favor of the male group, except for Var$_{95}$ and CVar$_{95}$ in GCD. This could be because when the confidence level ($\alpha$) is high, the tail-end distribution of costs across the two gender groups is similar.

\begin{table}[!htp]
    \fontsize{9}{12}\selectfont
    \centering
    \begin{tabular}{l|cc}
        \hline
         \multicolumn{3}{c}{\textbf{Adult Income}}\\
         \hline
         \makecell{Action\\Space} & \makecell{Action Costs\\$r(a)$} & \makecell{Transitions \\(Prob. of Success) $p(s')$}\\
         \hline
         Impr-Edu       & 2.0 & HS:0.9, Bachelors:0.8, \\ & &  Masters:0.9, Doc. 0.7 \\
         Incr-Hrs       & 0.5 & 1.0 \\
         Work-Gov       & 1.2 & 0.6 \\
         Work-Self      & 2.0 & 0.9 \\
         Work-Private   & 1.0 & 0.8 \\
         Job-Prof       & 1.0 & 0.4 \\
         Job-WhiteCol   & 1.0 & 0.7 \\
         \hline
         \hline
         \multicolumn{3}{c}{\textbf{German Credit}}\\
         \hline
        Impr-Skill      & 2.0 & Skilled:0.7, Manage:0.7 \\ 
        Incr-Savings    & 1.0 & 0.9 \\
        Incr-Checkings  & 1.0 & 0.9 \\
        Incr-Duration   & 2.0 & 1.0 \\
        House-Free      & 2.0 & 0.9 \\
        House-Rent      & 1.5 & 0.6 \\
        House-Own       & 1.0 & 0.3 \\
         \hline
    \end{tabular}
        \caption{Action Cost and Transition Models; If the action fails, the state stays the same.}
\label{tb:trans_cost_detail_env2}
\end{table}

\subsection{Parametric Analysis: Risk-Aversion $\beta$ and Horizon $H$}
We vary two parameters used in G-RSVI, risk-aversion level $\beta=0,0.25, 0.5, 1.0, 2.0$ and horizon $H=12, 8, 4$, and evaluate the performance on each of the parameter combinations in Table \ref{tb:eval_beta_h}. We use the full GCD dataset and 2000 randomly selected instances $n=2000$ in AID for evaluation. We also report the time spent for computing the policies.\footnote{The experiment is conducted on a MacOS with Apple M1 Max chip and RAM 32GB.} The measures again indicate better risk measures in risk-averse policies albeit with higher mean cost. The success rate significantly drops when the risk-aversion level $\beta=2.0$ and when horizon $H=4$ for AID (shaded in red). This happens because the emphasis on cost-variance is so strong, that it dwarfs the effect of the mean-cost, which in turn prevents the policy from getting to the favorable goal states within the finite horizon. The success rate and other risk measures for \textit{each} datapoint are evaluated using data from 100 Monte Carlo rollouts by following a policy. The $\rho_H$ estimates the success rate in the 100 rollouts. Empirically, $\rho_H$ is expected to be $<1.0$ with the presence of uncertainties.

\subsection{Health Insurance Premiums Dataset}
\label{apd:HIPD}
We also experimented on a Health Insurance Premiums Dataset (HIPD) from Kaggle \cite{health_dataset}. The dataset overview after pre-processing is shown in Table \ref{tb:HIPD_overview}. The cost and transition models are included in Table \ref{tb:trans_cost_detail_HIPD}. The evaluation in Table \ref{tb:eval_health_dataset} also shows better risk measures for the risk-averse policies computed, compared to the risk-neutral policy. The success rate is low because only three features are actionable: exercising, quitting smoking, and moving to a different region.

\clearpage
\begin{table}[!htp]
    \fontsize{10}{12}\selectfont
    \centering
    \begin{tabular}{c|l}
    \hline
        \multicolumn{2}{c}{\textbf{HIPD Feature Values }}\\
        \multicolumn{2}{c}{\textbf{Target: Charges=\{0,1\}}}\\
        \multicolumn{2}{c}{\textbf{\#States: 3456 \hfill Accuracy: 0.90 (Rand.Forest)}}\\
        \hline
        BMI    & $<$20, $<$25, $<$30, $<$35, $<$40, Above \\
        Smoker     & Yes, No \\
        Region & SW, SE, NW, NE\\
         \textbf{Children\#}    & 0, 1, 2, 3, 4 \\
        \textbf{Age}         & Young, Twenties, Twenties, \\
        & Thirties, Forties, Fifties, Older \\
         \textbf{Gender}     &  Female, Male \\
         \hline
    \end{tabular}
        \caption{The feature list and discrete values for HIPD. Bolded features are immutable. The other features are mutable and actionable.}
\label{tb:HIPD_overview}
\end{table}

\begin{table}[!htp]
    \fontsize{10}{12}\selectfont
    \centering
    \begin{tabular}{l|cc}
    \hline
         \multicolumn{3}{c}{\textbf{Health Insurance Premiums}}\\
         \hline
         \makecell{Action\\Space} & \makecell{Action Costs\\$r(a)$} & \makecell{Transitions\\(Prob. of Success) $p(s')$}\\
         \hline
         Exercise       & 0.8 & 0.95 \\
         QuitSmoke      & 0.8 & 0.5 \\
         Move-SW        & 1.0 & 0.7 \\
         Move-SE        & 1.0 & 0.7 \\
         Move-NW        & 1.0 & 0.7 \\
         Move-NE        & 1.0 & 0.7 \\
         \hline
    \end{tabular}
        \caption{Action Cost and Transition Models; If the action fails, the state stays the same.}
\label{tb:trans_cost_detail_HIPD}
\end{table}

\subsection{Lower Partial Standard Deviation}
\label{apd:LPSD}
G-RSVI penalizes the standard deviation $\sigma$ of the state values as stated in Equation \ref{eq:mean-variance Q-value}. Alternatively, we can penalize the Lower Partial Standard Deviation \cite{LPSD} which helps avoid situations requiring higher-than-average costs. In this case, the action value becomes:
\begin{equation*}
    Q_h(s,a) = \mu-\beta\sigma_{LP}[r(\cdot)+V_{h+1}(s')],
\end{equation*}
where $\mu=\mathbb{E}_{s'}[r(\cdot)+V_{h+1}(s')]$ is the average value, and
\begin{align*}
\sigma_{LP} & =\sqrt{\sum_{s'}p(s')[r(\cdot)+V_{h+1}(s')-\mu]^2}, \\ & \forall s' : r(\cdot)+V_{h+1}(s') < \mu.
\end{align*}
The evaluation on AID and GCD are included in Table \ref{tb:dataset_lpsd}, similar trends are observed as the results from Table \ref{tb:dataset_risk_measures}.

\clearpage

\begin{table*}[!htp]
    \fontsize{10}{12}\selectfont
    \centering
    \begin{tabular}{p{0.14\textwidth}|l|c|c|cc|cc|c|c}
         \hline
         \multicolumn{1}{c}{\textbf{Dataset}} &
         \multicolumn{1}{c}{\textbf{Policy}} & \multicolumn{1}{c}{$\pmb{\rho_{H=12}}$}& \multicolumn{1}{c}{$\pmb{(\mu_{cost},\sigma^2_{cost})}$}
         &
         \multicolumn{1}{c}{\textbf{VaR}{$\pmb{_{80}}$}}&
         \multicolumn{1}{c}{\textbf{CVaR}{$\pmb{_{80}}$}}&
         \multicolumn{1}{c}{\textbf{VaR}{$\pmb{_{95}}$}}&
         \multicolumn{1}{c}{\textbf{CVaR}{$\pmb{_{95}}$}}&
         \multicolumn{1}{c}{\textbf{Spars.}}&
         \multicolumn{1}{c}{\textbf{Proxi.}}\\
         \hline
        \multirow{4}{2.5cm}{\makecell{\textbf{Adult Income}\\($n=100$) \\ (Age $<30$)}}
        &DiCE10$_{\beta=0}$ & 0.774   & (5.70, 1.78)            & 5.14           & 7.48           & 6.21          & 8.64          & \textbf{2.42} & 4.94\\
        &$\beta=0$          & 0.997   & (\textbf{4.89}, 1.33)   & 5.19           & 7.15           & \textbf{6.00} & \textbf{8.10} & 2.75          & \textbf{4.37}\\
        &$\beta=0.25$       & 0.998   & (4.91, 1.08)            & 5.06           & \textbf{7.07}  & 6.03          & 8.18          & 2.80          & 4.56\\
        &$\beta=0.5$        & 0.995   & (4.95, \textbf{1.01})   & \textbf{5.04}  & 7.11           & 6.05          & 8.15          & 2.85          & 4.68\\
        \hline
        \multirow{4}{2.5cm}{\makecell{\textbf{German Credit}\\($n=100$)}}         
        &DiCE10$_{\beta=0}$ & 0.980   & (2.22, 0.78)            & 2.71           & 4.38           & 3.67          & 5.58          & \textbf{1.39}  & \textbf{1.63}\\
        &$\beta=0$          & 1.000   & (\textbf{1.53}, 0.34)   & 1.78           & 3.56           & 2.29          & \textbf{4.29} & \textbf{1.21} & \textbf{1.27}\\
        &$\beta=0.25$       & 1.000   & (1.55, 0.29)            & 1.73           & 3.51           & 2.29          & 4.39          & 1.23          & 1.32\\
        &$\beta=0.5$        & 1.000   & (1.56, \textbf{0.25})   & \textbf{1.72}  & \textbf{3.47}  & \textbf{2.23} & 4.30          & 1.30          & 1.39\\
         \hline
    \end{tabular}
    \caption{Compare risk-aware policies to the policy computed by using the final CEs provided by DiCE; We randomly select $n=100$ individual instances from AID and GCD; the same evaluation procedures are followed as Table \ref{tb:dataset_risk_measures}.}
\label{tb:dataset_risk_measures_100}
\end{table*}

\begin{table*}[ht]
    \fontsize{9}{12}\selectfont
    \centering
    \begin{tabular}{c|l|ccc|ccc}
         \hline
         \multicolumn{1}{c}{\textbf{Dataset}} &
         \multicolumn{1}{c}{\textbf{Policy}} &
         \multicolumn{1}{c}{\textbf{Female} $\pmb{\mu_{cost}}$} &
         \multicolumn{1}{c}{\textbf{Male} $\pmb{\mu_{cost}}$} &
         \multicolumn{1}{c}{\textbf{p-value}} &
        \multicolumn{1}{c}{\textbf{Female} $\pmb{\sigma^2_{cost}}$} &
         \multicolumn{1}{c}{\textbf{Male} $\pmb{\sigma^2_{cost}}$} &
         \multicolumn{1}{c}{\textbf{p-value}}\\
         \hline
        \multirow{3}{*}{\makecell{\textbf{Adult Income}\\(Female, $n=9824$)\\(Male, $n=16099$)}}
        &$\beta=0$    & 4.56  & 2.84  & $<$0.0001 & 1.41   & 1.12  & $<$0.0001\\
        &$\beta=0.25$ & 4.57  & 2.87  & $<$0.0001 & 1.18   & 0.72  & $<$0.0001\\
        &$\beta=0.5$  & 4.61  & 2.98  & $<$0.0001 & 1.11   & 0.57  & $<$0.0001\\
        \hline
        \multirow{3}{*}{\makecell{\textbf{German Credit}\\(Female, $n=103$)\\(Male, $n=178$)}}
        &$\beta=0$    & 1.72   & 1.61  & 0.3531 & 0.56   & 0.43  & 0.1688\\
        &$\beta=0.25$ & 1.75   & 1.62  & 0.2783 & 0.38   & 0.31  & 0.0964\\
        &$\beta=0.5$  & 1.79   & 1.65  & 0.3346 & 0.33   & 0.29  & 0.1637\\
         \hline
    \end{tabular}
    \caption{Mann-Whitney U test results for cost-mean and cost-variance between gender groups for AID and GCD}
\label{tb:gender_p_values}
\end{table*}

\begin{table*}[ht]
    \fontsize{9}{12}\selectfont
    \centering
    \begin{tabular}{c|l|ccc|ccc}
         \hline
         \multicolumn{1}{c}{\textbf{Dataset}} &
         \multicolumn{1}{c}{\textbf{Policy}} &
         \multicolumn{1}{c}{\textbf{Female} \textbf{VaR}$\pmb{_{80}}$} &
         \multicolumn{1}{c}{\textbf{Male} \textbf{VaR}$\pmb{_{80}}$} &
         \multicolumn{1}{c}{\textbf{p-value}} &
        \multicolumn{1}{c}{\textbf{Female} \textbf{CVaR}$\pmb{_{80}}$} &
         \multicolumn{1}{c}{\textbf{Male} \textbf{CVaR}$\pmb{_{80}}$} &
         \multicolumn{1}{c}{\textbf{p-value}}\\
         \hline
        \multirow{3}{*}{\makecell{\textbf{Adult Income}\\(Female, $n=9824$)\\(Male, $n=16099$)}}
        &$\beta=0$    & 4.76   & 3.27  & $<$0.0001 & 7.07   & 5.79  & $<$0.0001\\
        &$\beta=0.25$ & 4.64   & 3.08  & $<$0.0001 & 6.95   & 5.51  & $<$0.0001\\
        &$\beta=0.5$  & 4.66   & 3.10  & $<$0.0001 & 6.99   & 5.48  & $<$0.0001\\
        \hline
        \multirow{3}{*}{\makecell{\textbf{German Credit}\\(Female, $n=103$)\\(Male, $n=178$)}}
        &$\beta=0$    & 2.08   & 1.89  & 0.1678 & 3.70   & 3.64  & 0.9289\\
        &$\beta=0.25$ & 2.00   & 1.81  & 0.1934 & 3.61   & 3.50  & 0.5978\\
        &$\beta=0.5$  & 2.02   & 1.84  & 0.1677 & 3.64   & 3.50  & 0.6674\\
         \hline
    \end{tabular}
    \caption{Mann-Whitney U test results for VaR$_{80}$ and CVaR$_{80}$ between gender groups for AID and GCD}
\label{tb:gender_p_values_var_cvar_80}
\end{table*}

\begin{table*}[ht]
    \fontsize{9}{12}\selectfont
    \centering
    \begin{tabular}{c|l|ccc|ccc}
         \hline
         \multicolumn{1}{c}{\textbf{Dataset}} &
         \multicolumn{1}{c}{\textbf{Policy}} &
         \multicolumn{1}{c}{\textbf{Female} \textbf{VaR}$\pmb{_{95}}$} &
         \multicolumn{1}{c}{\textbf{Male} \textbf{VaR}$\pmb{_{95}}$} &
         \multicolumn{1}{c}{\textbf{p-value}} &
        \multicolumn{1}{c}{\textbf{Female} \textbf{CVaR}$\pmb{_{95}}$} &
         \multicolumn{1}{c}{\textbf{Male} \textbf{CVaR}$\pmb{_{95}}$} &
         \multicolumn{1}{c}{\textbf{p-value}}\\
         \hline
        \multirow{3}{*}{\makecell{\textbf{Adult Income}\\(Female, $n=9824$)\\(Male, $n=16099$)}}
        &$\beta=0$    & 5.70   & 4.30  & $<$0.0001 & 8.00   & 7.26  & $<$0.0001\\
        &$\beta=0.25$ & 5.48   & 3.95  & $<$0.0001 & 7.84   & 7.15  & $<$0.0001\\
        &$\beta=0.5$  & 5.16   & 3.92  & $<$0.0001 & 7.89   & 7.08  & $<$0.0001\\
        \hline
        \multirow{3}{*}{\makecell{\textbf{German Credit}\\(Female, $n=103$)\\(Male, $n=178$)}}
        &$\beta=0$    & 2.84   & 2.61  & 0.3029 & 4.66   & 4.62  & 0.7116\\
        &$\beta=0.25$ & 2.67   & 2.41  & 0.1353 & 4.63   & 4.42  & 0.3763\\
        &$\beta=0.5$  & 2.59   & 2.33  & 0.1814 & 4.51   & 4.29  & 0.5426\\
         \hline
    \end{tabular}
    \caption{Mann-Whitney U test results for VaR$_{95}$ and CVaR$_{95}$ between gender groups for AID and GCD}
\label{tb:gender_p_values_var_cvar_95}
\end{table*}
\begin{table*}[ht]
    \fontsize{10}{12}\selectfont
    \centering
    \begin{tabular}{c|l|c|c|cc|cc|c|c}
         \hline
         \multicolumn{1}{c}{\textbf{Dataset}} &
         \multicolumn{1}{c}{\textbf{Policy}} & \multicolumn{1}{c}{$\pmb{\rho_{H=12}}$}& \multicolumn{1}{c}{$\pmb{(\mu_{cost},\sigma^2_{cost})}$}
         &
         \multicolumn{1}{c}{\textbf{VaR}{$\pmb{_{80}}$}}&
         \multicolumn{1}{c}{\textbf{CVaR}{$\pmb{_{80}}$}}&
         \multicolumn{1}{c}{\textbf{VaR}{$\pmb{_{95}}$}}&
         \multicolumn{1}{c}{\textbf{CVaR}{$\pmb{_{95}}$}}&
         \multicolumn{1}{c}{\textbf{Spars.}}&
         \multicolumn{1}{c}{\textbf{Proxi.}}\\
         \hline
        \multirow{6}{*}{\makecell{\textbf{Adult Income}\\(Female, $n=1000$)\\~\\~\\\vspace{-3mm}\\(Male, $n=1000$)}}
        &$\beta=0$     & 0.989 & (\textbf{4.51}, 1.71)   & 4.75           & 7.01           & 5.53          & 7.86          & \textbf{2.57}  & \textbf{3.82}\\
        &$\beta=0.25$  & 0.986 & (\textbf{4.51}, 1.61)   & \textbf{4.71}  & \textbf{7.00}  & \textbf{5.45} & \textbf{7.83} & 2.59           & 3.85\\
        &$\beta=0.5$   & 0.977 & (4.56, \textbf{1.43})   & 4.81           & 7.07           & 5.46          & \textbf{7.83} & 2.61           & 3.93\\
        \hhline{|~|---------|}
        &$\beta=0$     & 0.997 & {(\textbf{2.67}, 0.96)}  & 2.97              & {\textbf{5.58}}   & 3.86               & {\textbf{6.69}}    & \textbf{1.95} & \textbf{2.51}\\
        &$\beta=0.25$  & 0.998 & (2.71, 0.81)            & {\textbf{2.96}}    & {\textbf{5.58}}    & 3.86                & 6.74             & 1.97          & 2.63\\
        &$\beta=0.5$   & 0.999 & {(2.79, \textbf{0.69})}   & 2.98             & 5.65             & {\textbf{3.78}}   & 6.81            & 2.05          & 2.73\\
        \hline
        \hline
        \multirow{6}{*}{\makecell{\textbf{German Credit}\\(Female, $n=103$)\\~\\~\\\vspace{-3mm}\\(Male, $n=178$)}}         
        &$\beta=0$     & 0.999 & (\textbf{1.65}, 0.32)   & 1.69           & 2.85           & {\textbf{2.49}} & {\textbf{3.84}}  & \textbf{1.22} & \textbf{1.38}\\
        &$\beta=0.25$  & 1.000 & (1.67, 0.26)            & 1.68           & 2.84           & 2.59               & 4.03                & 1.24          & 1.41\\
        &$\beta=0.5$   & 1.000 & (1.67, \textbf{0.24})   & \textbf{1.66}  & \textbf{2.80}  & 2.57               & 4.00                & 1.24          & 1.42\\
        \hhline{|~|---------|}
        &$\beta=0$     & 1.000 & (\textbf{1.62}, 0.26)      & 1.64              & II2.79             & 2.51                 & 3.87               & \textbf{1.20} & \textbf{1.36}\\
        &$\beta=0.25$  & 1.000 & (1.63, 0.23)              & 1.62                & 2.77               & 2.46            & 3.79             & 1.21          & 1.38\\
        &$\beta=0.5$   & 1.000 & {(1.63, \textbf{0.22})}  & {\textbf{1.61}}   & {\textbf{2.74}}  & {\textbf{2.42}} & {\textbf{3.62}} & 1.22          & 1.39\\
         \hline
    \end{tabular}
    \caption{Evaluating recourse policies across gender using a different set of cost and transition functions; the same evaluation procedures are followed as Table \ref{tb:dataset_risk_measures}.}
\label{tb:env2_risk_measures_gender}
\end{table*}

\begin{table*}[t]
    \fontsize{10}{12}\selectfont
    \centering
    \begin{tabular}{c|l|c|c|cc|cc|c|c}
         \hline
         \multicolumn{1}{c}{\textbf{Dataset}} &
         \multicolumn{1}{c}{\textbf{Policy}} & \multicolumn{1}{c}{$\pmb{\rho_{H}}$}& \multicolumn{1}{c}{$\pmb{(\mu_{cost},\sigma^2_{cost})}$}
         &
         \multicolumn{1}{c}{\textbf{VaR}{$\pmb{_{80}}$}}&
         \multicolumn{1}{c}{\textbf{CVaR}{$\pmb{_{80}}$}}&
         \multicolumn{1}{c}{\textbf{VaR}{$\pmb{_{95}}$}}&
         \multicolumn{1}{c}{\textbf{CVaR}{$\pmb{_{95}}$}}&
         \multicolumn{1}{c}{\textbf{Spars.}}&
         \multicolumn{1}{c}{\textbf{Proxi.}}\\
         \hline
        \multirow{5}{2.5cm}{\makecell{\textbf{Adult Income}\\($n=2000$)\\Horizon $=12$\\Time: 23.12 (s)\\~}}         
        &$\beta=0$    & 0.995           & (\textbf{3.69}, 1.26)   & 3.97           & 6.46           & 4.87          & 7.57          & \textbf{2.20} & \textbf{3.05}\\
        &$\beta=0.25$ & 0.995           & (3.71, 0.93)            & 3.79           & 6.26           & 4.63          & 7.49          & 2.25          & 3.23\\
        &$\beta=0.5$  & 0.994           & (3.78, 0.83)            & 3.82           & 6.28           & 4.62          & 7.49          & 2.29          & 3.34\\
        &$\beta=1.0$  & 0.978           & (4.29, 0.75)            & 4.12           & 6.96           & 4.69          & 8.05          & 2.38          & 3.78\\
        &$\beta=2.0$  & 0.717     & (3.82, \textbf{0.28})   & \textbf{3.34}  & \textbf{5.72}  & \textbf{4.06} & \textbf{7.24} & 2.28          & 3.36\\
        \hline
        \multirow{5}{2.5cm}{\makecell{~\\~\\Horizon $=8$\\Time: 15.89 (s)\\~}}         
        &$\beta=0$    & 0.915           & (3.30, 0.83)                      & 3.77           & 5.85           & 4.80          & 6.74          & \textbf{2.14} & \textbf{2.77}\\
        &$\beta=0.25$ & 0.905           & (3.29, 0.60)                      & 3.57           & 5.63           & 4.55          & 6.71          & 2.21          & 2.89\\
        &$\beta=0.5$  & 0.891           & (3.28, 0.56)                      & 3.52           & 5.60           & 4.51          & 6.74          & 2.24          & 2.90\\
        &$\beta=1.0$  & 0.852           & (3.48, 0.43)                      & 3.60           & 5.75           & 4.52          & 6.89          & 2.33          & 3.13\\
        &$\beta=2.0$  & 0.611     & (\textbf{3.13}, \textbf{0.16})    & \textbf{3.10}  & \textbf{5.15}  & \textbf{3.73} & \textbf{5.15} & 2.19          & 2.86\\
        \hline
        \multirow{5}{2.5cm}{\makecell{~\\~\\Horizon $=4$\\Time: 7.90 (s)\\~}}       
        &$\beta=0$    & 0.544     & (\textbf{1.97}, 0.19)             & 2.22           & 3.14           & 2.54          & 3.74          & \textbf{1.69} & \textbf{1.79}\\
        &$\beta=0.25$ & 0.548     & (2.00, 0.15)                      & 2.20           & \textbf{3.10}  & 2.50          & \textbf{3.55} & 1.76          & 1.87\\
        &$\beta=0.5$  & 0.552     & (2.05, 0.12)                      & 2.19           & 3.22           & 2.51          & 3.77          & 1.83          & 1.95\\
        &$\beta=1.0$  & 0.517     & (2.09, 0.09)                      & 2.19           & 3.39           & 2.47          & 3.78          & 1.88          & 2.02\\
        &$\beta=2.0$  & 0.335     & (2.13, \textbf{0.03})             & \textbf{2.12}  & 3.27           & \textbf{2.35} & 3.96          & 1.73          & 2.08\\
        \hline
        \hline
        \multirow{5}{2.5cm}{\makecell{\textbf{German Credit}\\($n=281$)\\Horizon$=12$\\Time: 60.02 (s)\\~}}         
        &$\beta=0$    & 1.000 & (\textbf{1.65}, 0.48)   & 1.96           & 3.66           & 2.63          & 4.56          & \textbf{1.26} & \textbf{1.33}\\
        &$\beta=0.25$ & 1.000 & (1.67, 0.35)            & \textbf{1.87}  & \textbf{3.51}  & 2.51          & 4.50          & 1.34          & 1.43\\
        &$\beta=0.5$  & 1.000 & (1.70, 0.30)            & 1.90           & 3.56           & 2.48          & \textbf{4.40} & 1.40          & 1.50\\
        &$\beta=1.0$  & 1.000 & (1.93, \textbf{0.25})   & 2.04           & 3.90           & 2.57          & 4.74          & 1.62          & 1.79\\
        &$\beta=2.0$  & 0.981 & (2.88, 0.65)            & 2.03           & 4.44           & \textbf{2.38} & 4.53          & 1.82          & 2.08\\
         \hline
       \multirow{5}{2.5cm}{\makecell{~\\~\\Horizon$=8$\\Time: 40.45 (s)\\~}}         
        &$\beta=0$    & 0.998 & (\textbf{1.64}, 0.43)   & 1.94           & 3.58           & 2.65          & 4.45          & \textbf{1.26} & \textbf{1.33}\\
        &$\beta=0.25$ & 0.999 & (1.66, 0.33)            & \textbf{1.88}  & \textbf{3.51}  & 2.52          & 4.47          & 1.34          & 1.42\\
        &$\beta=0.5$  & 1.000 & (1.70, 0.29)            & 1.92           & 3.56           & \textbf{2.50} & \textbf{4.41} & 1.40          & 1.50\\
        &$\beta=1.0$  & 1.000 & (1.91, \textbf{0.23})   & 2.03           & 3.90           & 2.55          & 4.69          & 1.61          & 1.79\\
        &$\beta=2.0$  & 0.983 & (2.49, 0.31)            & 2.44           & 4.75           & 2.92          & 5.61          & 1.81          & 2.08\\
         \hline
        \multirow{5}{2.5cm}{\makecell{~\\~\\Horizon$=4$\\Time: 20.56 (s)\\~}}         
        &$\beta=0$    & 0.963 & (\textbf{1.54}, 0.22)   & 1.81           & \textbf{3.05}  & 2.37          & 3.42          & \textbf{1.26} & \textbf{1.32}\\
        &$\beta=0.25$ & 0.969 & (1.56, 0.19)            & \textbf{1.78}  & 3.06           & \textbf{2.24} & 3.40          & 1.31          & 1.37\\
        &$\beta=0.5$  & 0.972 & (1.59, 0.18)            & 1.81           & 3.06           & 2.25          & \textbf{3.37} & 1.35          & 1.42\\
        &$\beta=1.0$  & 0.975 & (1.69, 0.16)            & 1.86           & 3.24           & 2.25          & 3.75          & 1.51          & 1.59\\
        &$\beta=2.0$  & 0.925 & (2.18, \textbf{0.07})   & 2.24           & 3.68           & 2.51          & 3.91          & 1.71          & 1.99\\
         \hline
    \end{tabular}
    \caption{Parametric evaluation on risk-aversion $\beta$ and horizon $H$ for AID and GCD; the same evaluation procedures are followed as Table \ref{tb:dataset_risk_measures}. The time spent to compute the policies for all states (57600 in AID, 147456 in GCD) is included.}
\label{tb:eval_beta_h}
\end{table*}

\begin{table*}[!t]
    \fontsize{10}{12}\selectfont
    \centering
    \begin{tabular}{c|l|c|c|cc|cc|c|c}
         \hline
         \multicolumn{1}{c}{\textbf{Dataset}} &
         \multicolumn{1}{c}{\textbf{Policy}} & \multicolumn{1}{c}{$\pmb{\rho_{H=12}}$}& \multicolumn{1}{c}{$\pmb{(\mu_{cost},\sigma^2_{cost})}$}
         &
         \multicolumn{1}{c}{\textbf{VaR}{$\pmb{_{80}}$}}&
         \multicolumn{1}{c}{\textbf{CVaR}{$\pmb{_{80}}$}}&
         \multicolumn{1}{c}{\textbf{VaR}{$\pmb{_{95}}$}}&
         \multicolumn{1}{c}{\textbf{CVaR}{$\pmb{_{95}}$}}&
         \multicolumn{1}{c}{\textbf{Spars.}}&
         \multicolumn{1}{c}{\textbf{Proxi.}}\\
         \hline
        \multirow{4}{2.5cm}{\makecell{~\\\textbf{HIPD}\\($n=604$)\\~}}
        &$\beta=0$    & 0.717 & (1.74, 0.86)            & 2.29           & 3.62           & 3.42          & 4.84          & 1.21          & \textbf{1.38}\\
        &$\beta=0.25$ & 0.717 & (\textbf{1.73}, 0.85)   & \textbf{2.28}  & 3.62           & 3.41          & 4.82          & 1.21          & \textbf{1.38}\\
        &$\beta=0.5$  & 0.717 & (1.75, \textbf{0.82})   & 2.29           & \textbf{3.59}  & \textbf{3.36} & \textbf{4.80} & \textbf{1.20} & 1.43\\
        &$\beta=1.0$  & 0.716 & (2.79, 0.83)            & 3.29           & 4.58           & 4.42          & 5.88          & 1.66          & 2.68\\
        \hline
    \end{tabular}
    \caption{Evaluating recourse policies for Health Insurance Premiums Dataset; the same evaluation procedures are followed as Table \ref{tb:dataset_risk_measures}.}
\label{tb:eval_health_dataset}
\end{table*}
\begin{table*}[!t]
    \fontsize{10}{12}\selectfont
    \centering
    \begin{tabular}{c|l|c|c|cc|cc|c|c}
         \hline
         \multicolumn{1}{c}{\textbf{Dataset}} &
         \multicolumn{1}{c}{\textbf{Policy}} & \multicolumn{1}{c}{$\pmb{\rho_{H=12}}$}& \multicolumn{1}{c}{$\pmb{(\mu_{cost},\sigma^2_{cost})}$}
         &
         \multicolumn{1}{c}{\textbf{VaR}{$\pmb{_{80}}$}}&
         \multicolumn{1}{c}{\textbf{CVaR}{$\pmb{_{80}}$}}&
         \multicolumn{1}{c}{\textbf{VaR}{$\pmb{_{95}}$}}&
         \multicolumn{1}{c}{\textbf{CVaR}{$\pmb{_{95}}$}}&
         \multicolumn{1}{c}{\textbf{Spars.}}&
         \multicolumn{1}{c}{\textbf{Proxi.}}\\
         \hline
        \multirow{5}{2.5cm}{\makecell{\textbf{Adult Income}\\($n=25923$)\\~\\}}
        &$\beta=0$    & 0.994 & (\textbf{3.49}, 1.23)   & 3.81           & 6.31           & 4.76          & 7.53          & \textbf{2.09} & \textbf{2.87}\\
        &$\beta=0.25$ & 0.994 & (3.51, 0.91)            & \textbf{3.65}  & \textbf{6.10}  & 4.46          & 7.44          & 2.17          & 3.05\\
        &$\beta=0.5$  & 0.994 & (3.53, 0.86)            & \textbf{3.65}  & 6.12           & \textbf{4.44} & \textbf{7.43} & 2.18          & 3.09\\
        &$\beta=0.75$ & 0.992 & (3.60, \textbf{0.76})   & 3.67           & 6.13           & 4.45          & \textbf{7.43} & 2.21          & 3.20\\

        \hline
        \multirow{6}{2.5cm}{\makecell{\textbf{German Credit}\\($n=281$)\\~\\~\\}}         
        &$\beta=0$    & 1.000 & (\textbf{1.65}, 0.48)   & 1.96           & 3.66           & 2.63          & 4.56          & \textbf{1.26} & \textbf{1.33}\\
        &$\beta=0.25$ & 1.000 & (1.66, 0.38)            & \textbf{1.88}  & \textbf{3.56}  & 2.53          & 4.52          & 1.33          & 1.41\\
        &$\beta=0.5$  & 1.000 & (1.68, 0.33)            & 1.91           & 3.56           & 2.50          & \textbf{4.46} & 1.36          & 1.45\\
        &$\beta=0.75$ & 1.000 & (1.74, \textbf{0.28})   & 1.93           & 3.72           & \textbf{2.47} & 4.73          & 1.50          & 1.59\\
         \hline
    \end{tabular}
    \caption{Evaluating recourse policies by penalizing LPSD with different risk-aversion levels $\beta$ and horizon $H=12$ for AID, GCD; the same evaluation procedures are followed as Table \ref{tb:dataset_risk_measures}.}
\label{tb:dataset_lpsd}
\end{table*}

\clearpage
\clearpage
\section{Synthetic Motivation Example}
\label{apd:syn_domain}
\begin{figure}[!htp]
    \centering
    \begin{subfigure}{0.47\textwidth}
        \centering
        \includegraphics[width=1\textwidth]{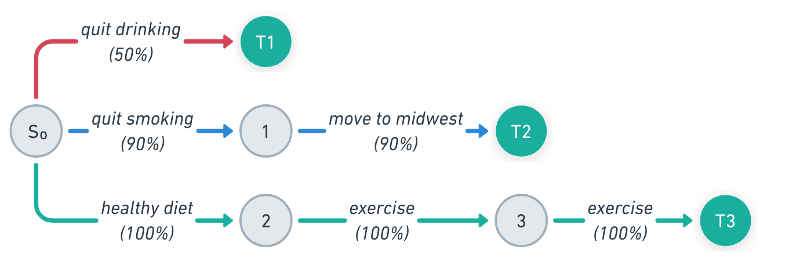}
        \caption{Transition probabilities}
        \label{fig:syn_dyn}
    \end{subfigure}
    \begin{subfigure}{0.47\textwidth}
        \centering
        \includegraphics[width=1\textwidth]{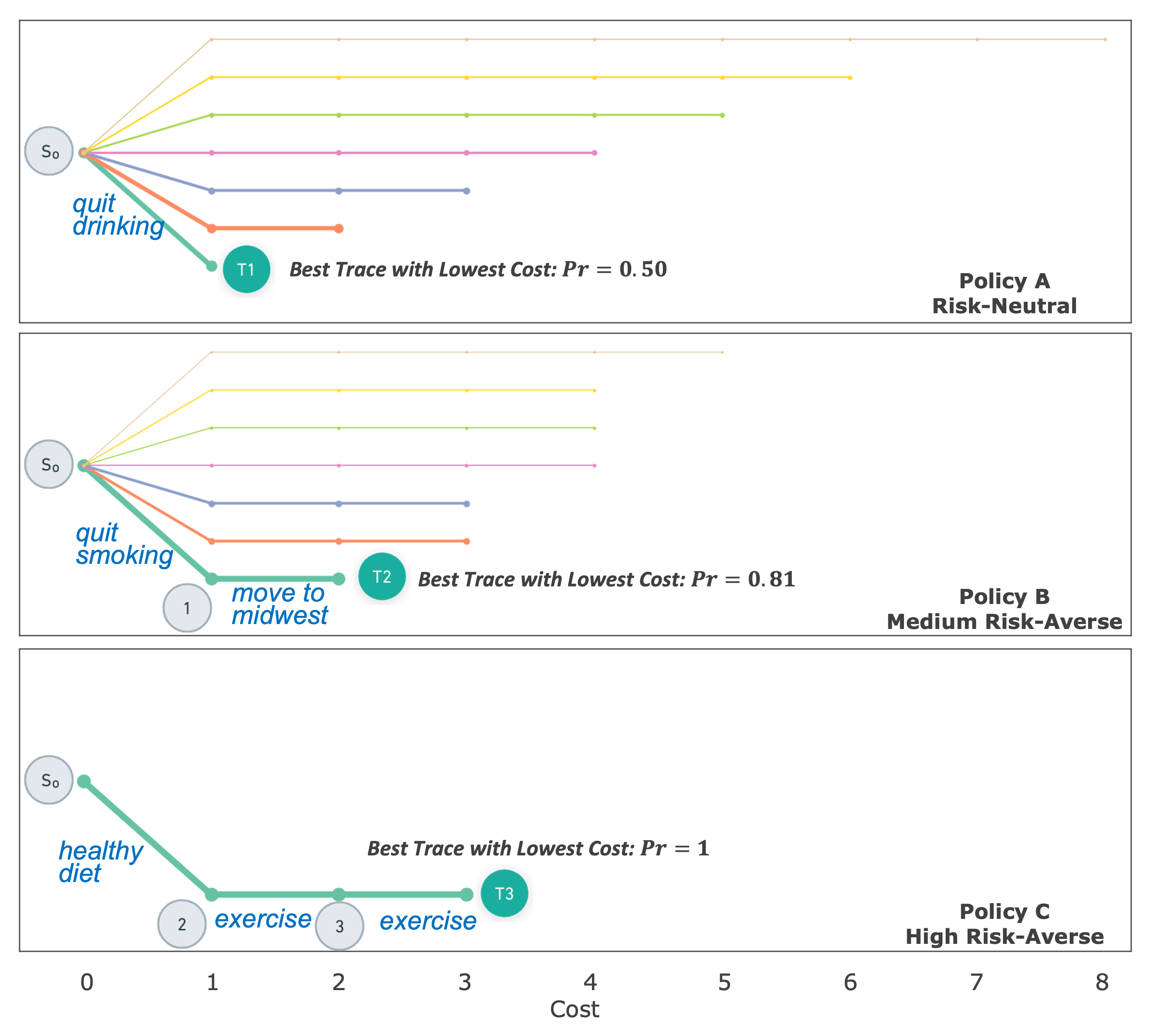}
        \caption{Recourse policy visualization}
        \label{fig:syn_policy_viz}
    \end{subfigure}
    \caption{Health insurance premiums recourse (synthetic)}  
    \label{fig:syn_result}
\end{figure}

We consider a synthetic motivational example for reducing health insurance premiums. This motivational example is different from the dataset in Section \ref{apd:HIPD}; this example  uses a simpler set of states and possible actions to illustrate difference policies and risk. For this example we define a model of actions (transition probabilities and costs) as well as a decision function $f$ that a heath insurance company uses to determine the insurance premium. In this example, consider that insurance premiums are charged at two levels (high, and low). Let us consider an individual with the following features $s_o=[\text{smoking, drinking, high cholesterol, high BMI, west}]$ is classified to receive the high insurance premium, which is unfavorable $f(s_o)=y^-$. Figure \ref{fig:syn_dyn} describes the possible actions the individual might take at each state with the transition (success) probabilities , where $T\in\{T1,T2,T3\}$ are the states with the desired outcomes $f(T)=y^+$ (low insurance premium). We assume the cost is the same per step (time cost of 3-months). Taking the \textit{quit-drinking} action at $s_o$ would lead to $T1$ directly (red path) but only with $50\%$ success rate in three months; taking the \textit{quit-smoking} and \textit{move-to-midwest} actions would follow the blue path, where each step has success rate $90\%$; the \textit{health-diet} and \textit{exercise}  actions would progress the feature state along the green path to reduce cholesterol and BMI with $100\%$ success rate but the policy would take 9 months. The feature state remains the same when the action fails.

We apply the G-RSVI algorithm with risk-aversion levels $\beta=0,0.5,\text{and}\ 1$ for risk-neutral, medium risk-averse, and high risk-averse policies respectively. The possible encountered state sequences (traces) for three polices are shown in Figure \ref{fig:syn_policy_viz}, and the lowest cost trace is highlighted with line thickness indicating the probability. Using 1000 rollouts of each policy, the risk measures of the three policies are summarized in Table \ref{tb:syn_risk_measures}. The success rate $\rho_{H=8}$ --which is success probability of recourse within 8 horizon steps-- can be less than $100\%$ because some actions have less than 100\% probability of success. When $\beta > 0$, the recourse policy achieves lower $\sigma^2_{cost}$ and lower cost in VaR$_{95}$ and CVaR$_{95}$, in which the individual has high confidence of a successful recourse with low cost-risk. CVaR$_{95}$ for $\beta=1$ is n/a because, $100\%$ of the time, recourse is achieved within cost 3. By changing the risk-aversion parameter $\beta$, we are able to provide a set of policies corresponding to different risk tolerance levels for the individual.

\begin{table}[!htp]
    \fontsize{10}{12}\selectfont
    \centering
    \begin{tabular}{l|c|c|cc}
         \hline
         \multicolumn{1}{c}{\textbf{Policy}} & \multicolumn{1}{c}{$\pmb{\rho_{H=8}}$}& \multicolumn{1}{c}{$\pmb{(\mu_{cost},\sigma^2_{cost})}$}
         &
         \multicolumn{1}{c}{\textbf{VaR}{$\pmb{_{95}}$}}&
         \multicolumn{1}{c}{\textbf{CVaR}{$\pmb{_{95}}$}}\\
         \hline
         $\beta=0$ & 0.999 & (\textbf{1.9}, 1.7) & 5.0 & 6.5\\
        $\beta=0.5$ & 1.000 & (2.2, 0.2) & 3.0 & 4.1\\
        $\beta=1.0$ & 1.000 & (3.0, \textbf{0.0}) & \textbf{3.0} & \textbf{n/a}\\
         \hline
    \end{tabular}
    \caption{Evaluating risk measures}
\label{tb:syn_risk_measures}
\end{table}

\clearpage
\section{Methodology}

\subsection{Greedy Risk-Sensitive Episodic Value Iteration (G-RSEVI)}
G-RSVI is suitable for fast policy computation for all possible recourse states. However, recourse is often tailored and offered to a single individual, and the majority of the states are not reachable by the individual depending on their initial state or current features. Therefore, instead of exploring the full state space, we introduce G-RSEVI in Algorithm \ref{alg:GRSEVI}, an episodic variant of G-RSVI that finds the policy for a certain initial state. We allow $E$ episodes for computing the policy (Line 2), and the recourse always starts at the initial state of the recipient (Line 4). For each episode, we select the action based on the current action value $Q_h(s,a)$ with $\epsilon$-greedy, and $\epsilon$ is decaying exponentially. We then collect the state-action pairs for $H$ steps (Line 5-8). With the stored transition data $D$, we update the action value $Q_h(s,a)$ and state value $V_h(s)$ with the risk-averse objective backward (Line 10-17). G-RSEVI ($\mathcal{O}(|E||S||A||H|)$) provides additional benefits in computation complexity when the state space is extremely high (i.e., when $|E|\ll |S|$), compared to G-RSVI ($\mathcal{O}(|S^2||A||H|)$).
\begin{algorithm}[h]
\caption{G-RSEVI}
\hangindent=10.5mm \textbf{Input:} recourse MDP$\langle S,A,T,R,H \rangle$, ML model $f$, initial feature state $s_o$

\hangindent=17.6mm \textbf{Parameters:} risk aversion level $\beta\in[0,\infty)$, max episode $E$

\begin{algorithmic}[1]
\label{alg:GRSEVI}
\STATE $V_{H+1}(s)\leftarrow 0, \forall s \in S$
\FOR{episode $k=1,\dots,E$}
\STATE $D\leftarrow empty$
\STATE $s_1 \leftarrow$ Reset state to $s_o$
\FOR{step $h=1,\dots,H$}
\STATE $a_h\leftarrow \argmax Q_h(s_h,\cdot)$ $\epsilon$-greedy action selection
\STATE $D\leftarrow$ store $(s_h,a_h)$
\STATE $s_{h+1} \leftarrow$ take action and receive the next state
\ENDFOR
\FOR{step $h=H,H-1,\dots,1$}
\STATE $s,a\leftarrow$ get $(s_h,a_h)$ from $D$
\STATE $r(s'),p(s')\leftarrow$ receive reward and transition
\STATE $\mu\leftarrow\sum_{s'}p(s')[r(s')+V_{h+1}(s')]$
\STATE $\sigma^2\leftarrow\sum_{s'}p(s')[r(s')+V_{h+1}(s')-\mu]^2$
\STATE $Q_{h}(s,a)\leftarrow \mu-\beta\sigma$ (Equation \ref{eq:mean-variance Q-value})
\STATE $V_{h}(s)\leftarrow \max Q_{h}(s,\cdot)$
\STATE $\pi_h(s)=\argmax Q_h(s,\cdot)$
\ENDFOR
\ENDFOR
\RETURN Recourse policy $\pi_h(s)$
\end{algorithmic}
\end{algorithm}

\begin{figure}[!ht]
    \centering
    \begin{subfigure}{0.47\textwidth}
        \centering
        \includegraphics[width=1\textwidth]{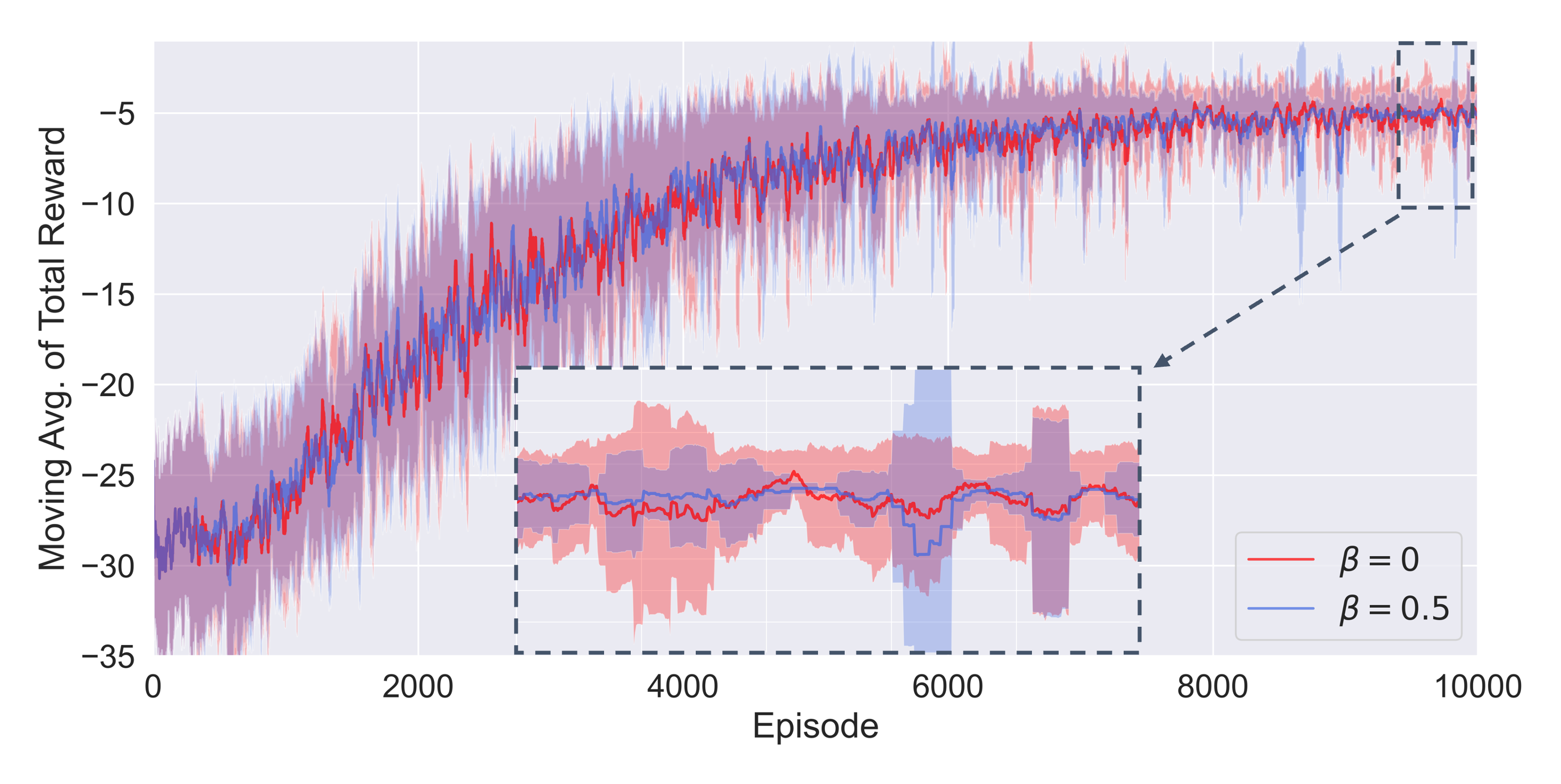}
        \caption{Example instance in AID}
        \label{fig:adult16_rsviep}
    \end{subfigure}
    \begin{subfigure}{0.47\textwidth}
        \centering
        \includegraphics[width=1\textwidth]{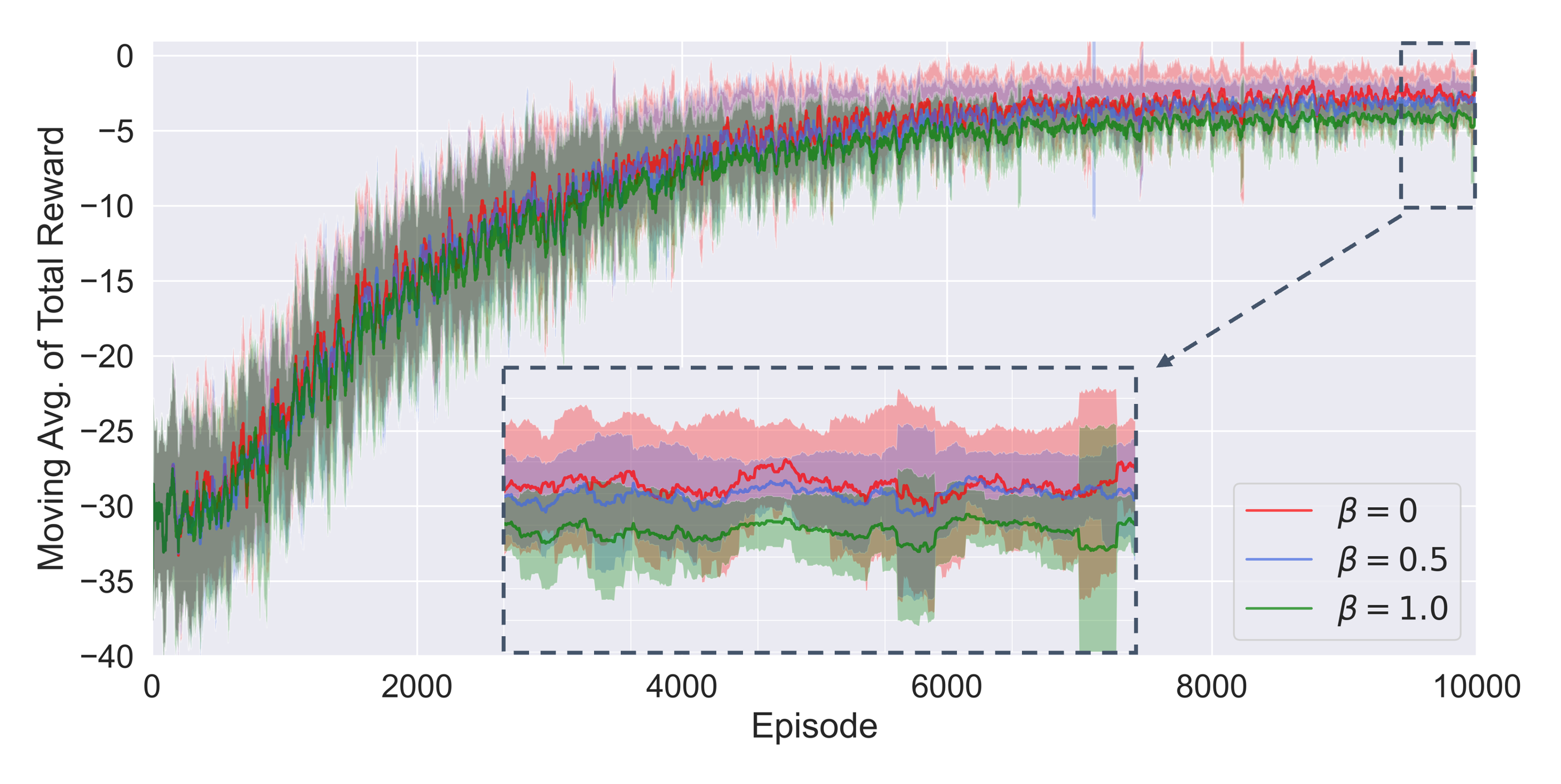}
        \caption{Example instance in GCD}
        \label{fig:credit44_rsviep}
    \end{subfigure}
    \caption{Moving average of total recourse cost during G-RSEVI updates with maximum episode 10,000 and exponential decay factor 0.9995 for $\epsilon$. Shades indicate $\pm\sigma$ cost range of the moving window.}  
    \label{fig:rsvi_episodic}
\end{figure}

We visualize the total recourse cost vs. episode in Figure \ref{fig:rsvi_episodic} for the two example instances selected from AID and GCD (same instances as in Table \ref{tb:dataset_risk_measures}). Towards the last 500 episodes (low chance of random actions), we show the variability in costs for higher risk-aversion policies is less than the risk-neutral ones.

\subsection{Case Study: Example Instances in Datasets}
\label{apd:instances}
\begin{figure*}[!htp]
    \centering
    \begin{subfigure}{0.47\textwidth}
        \centering
        \includegraphics[width=1\textwidth]{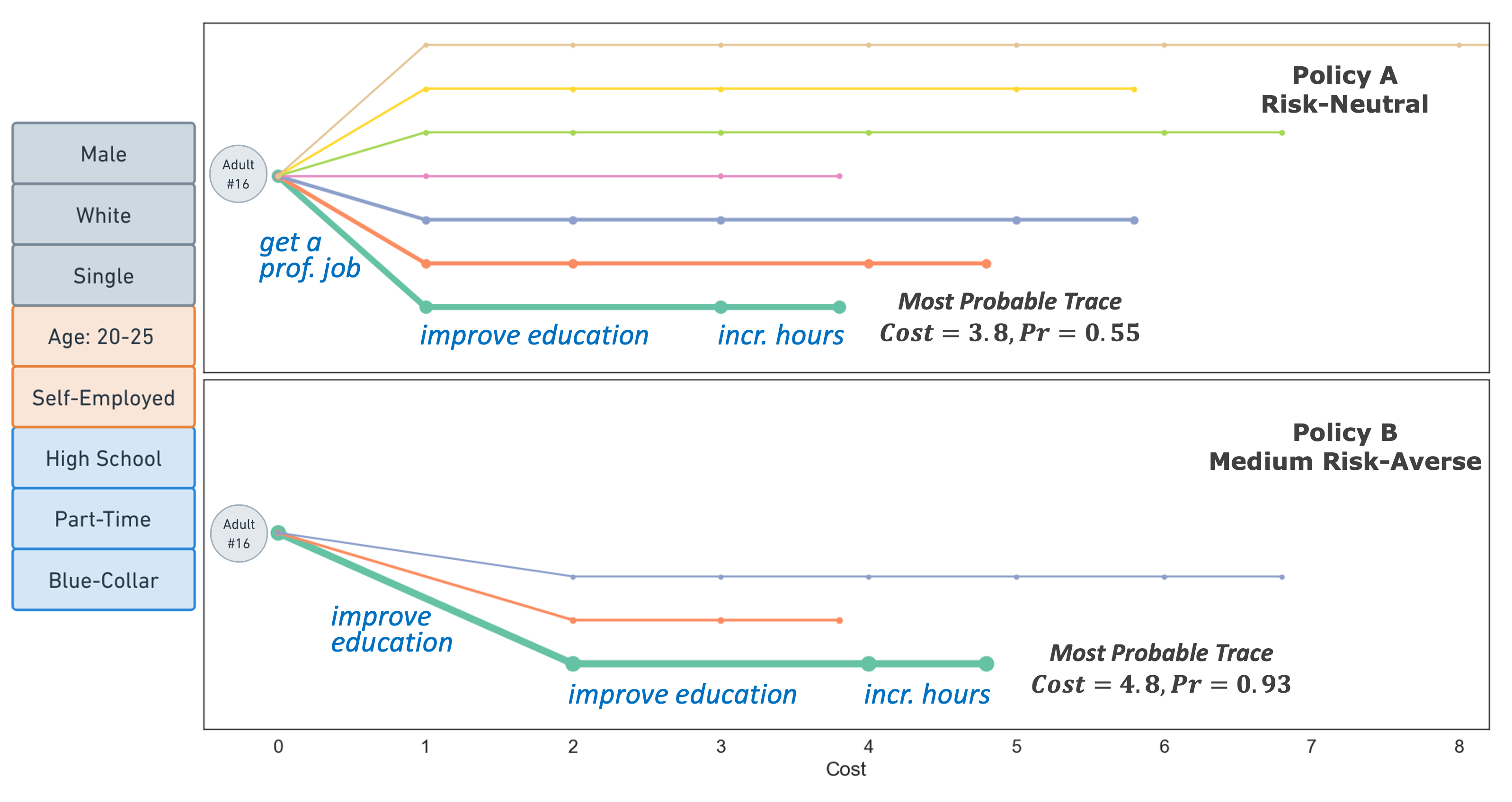}
        \caption{Instance \#16 in AID}
        \label{fig:adult16_policy_viz_apd}
    \end{subfigure}
    \hfill
    \begin{subfigure}{0.47\textwidth}
        \centering
        \includegraphics[width=1\textwidth]{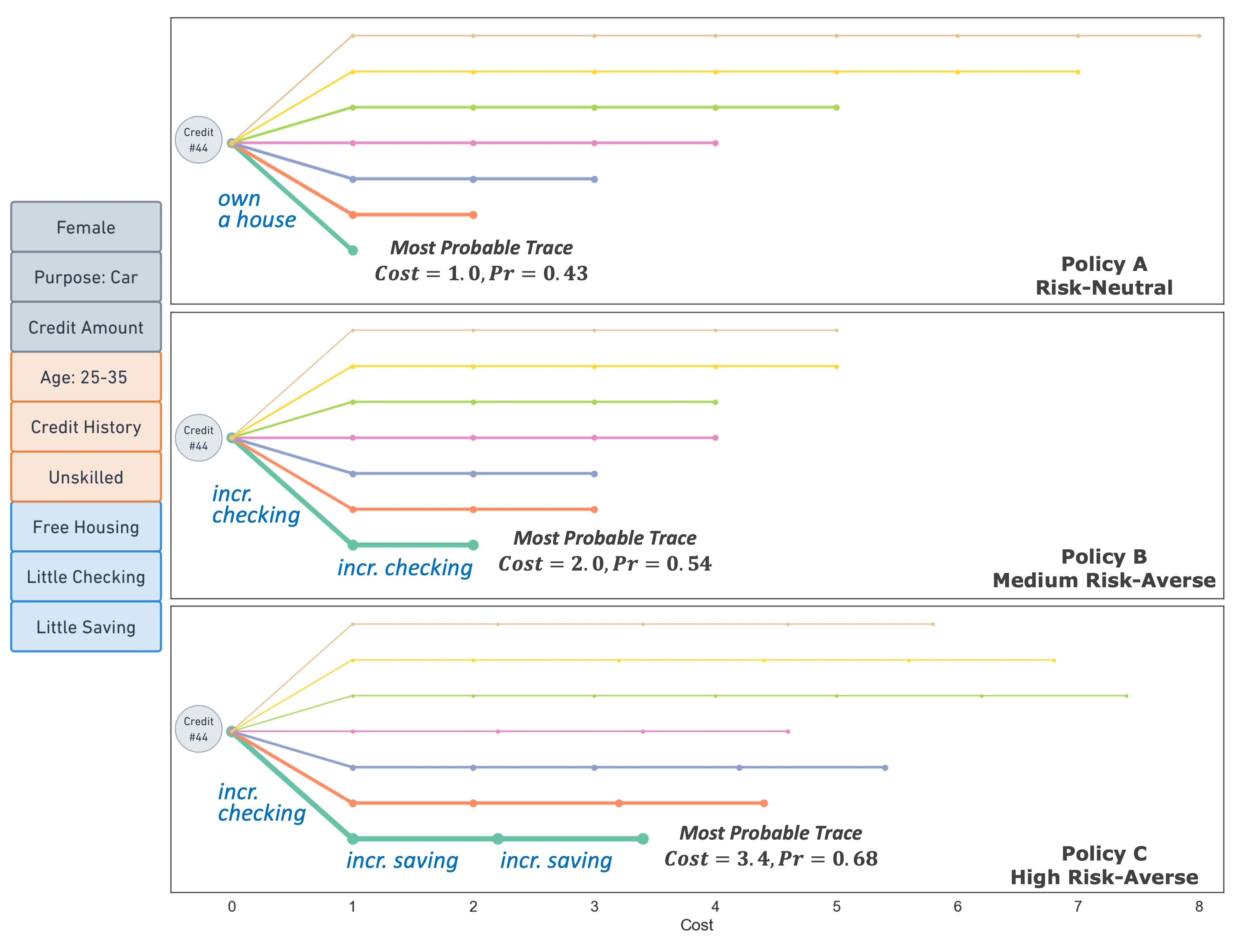}
        \caption{Instance \#44 in GCD}
        \label{fig:credit44_policy_viz_apd}
    \end{subfigure}
    \begin{subfigure}{0.47\textwidth}
        \centering
        \includegraphics[width=1\textwidth]{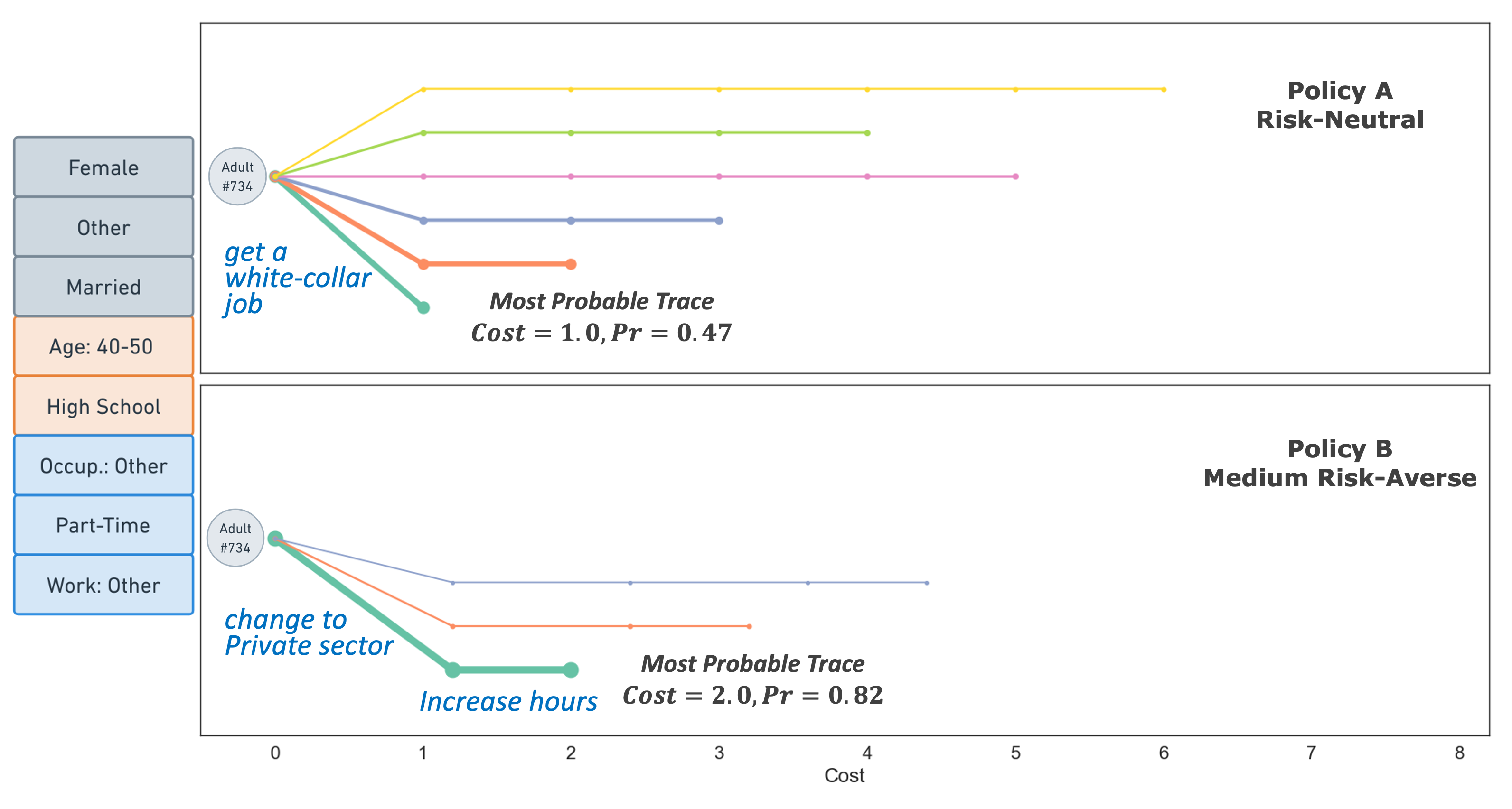}
        \caption{Instance \#734 in AID}
        \label{fig:adult734_policy_viz_apd}
    \end{subfigure}
    \hfill
    \begin{subfigure}{0.47\textwidth}
        \centering
        \includegraphics[width=1\textwidth]{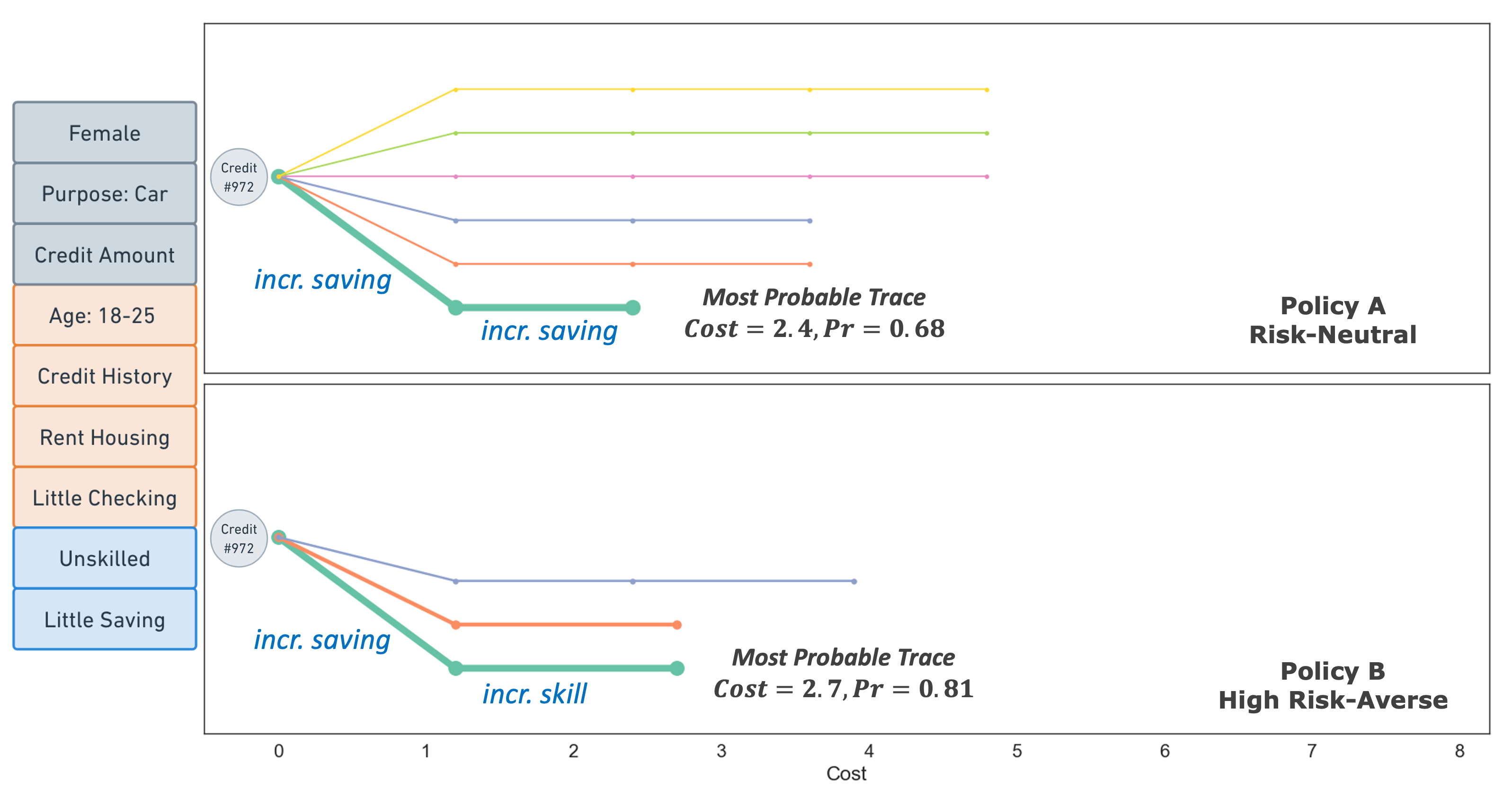}
        \caption{Instance \#972 in GCD}
        \label{fig:credit972_policy_viz_apd}
    \end{subfigure}
    \caption{Policy visualizations; $\beta=0,0.5,1.0$ correspond to risk-neutral, medium risk-averse, and high-risk-averse respectively.}  
    \label{fig:policy_viz_all}
\end{figure*}

In this section, we provide additional visualizations of policies computed by G-RSVI as shown in Figure \ref{fig:policy_viz_all}. Instance \#16 in AID and Instance \#44 in GCD correspond to the two example instances used in Table \ref{tb:dataset_risk_measures} as well as in Figure \ref{fig:rsvi_episodic}.

\end{document}